\documentclass{bmvc2k}

\usepackage{multirow}
\usepackage[table, dvipsnames]{xcolor}
\usepackage{mathrsfs}

\title{Photorealistic Style Transfer with Screened Poisson Equation}

\addauthor{Roey Mechrez}{http://cgm.technion.ac.il/people/Roey}{1}

\addauthor{Eli Shechtman}{research.adobe.com/person/eli-shechtman}{2}

\addauthor{Lihi Zelnik-Manor}{http://lihi.eew.technion.ac.il}{1}

\addinstitution{
Technion Israel Institute of Technology\\
Haifa, Israel\\
}
\addinstitution{
Adobe Research\\
Seattle, USA\\
}

\runninghead{Mechrez \bmvaEtAl}{Screened Poisson for Photorealistic Style Transfer}
\begin{document}

\maketitle

\begin{abstract}
Recent work has shown impressive success in transferring painterly style to images. These approaches, however, fall short of photorealistic style transfer. Even when both the input and reference images are photographs, the output still exhibits distortions reminiscent of a painting. 
In this paper we propose an approach that takes as input a stylized image and makes it more photorealistic. It relies on the Screened Poisson Equation, maintaining the fidelity of the stylized image while constraining the gradients to those of the original input image. Our method is fast, simple, fully automatic and shows positive progress in making a stylized image photorealistic. Our results exhibit finer details and are less prone to artifacts than the state-of-the-art.
\end{abstract}

%-------------------------------------------------------------------------
\section{Introduction}

\begin{figure}[t]
		\setlength{\tabcolsep}{.15em}
        \small
        \begin{tabular}{cccc} 
		\includegraphics[width=.24\linewidth]{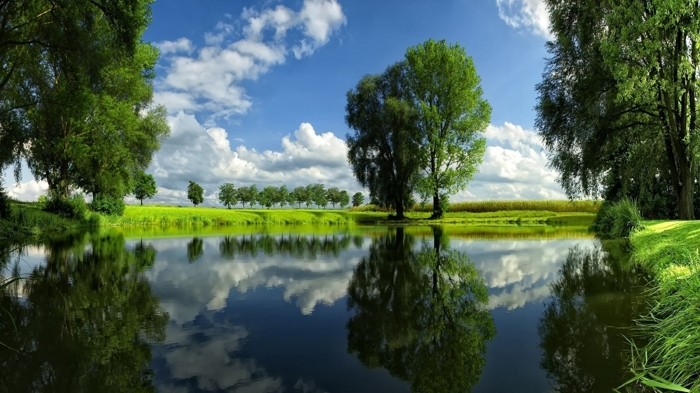}&
		\includegraphics[width=.24\linewidth]{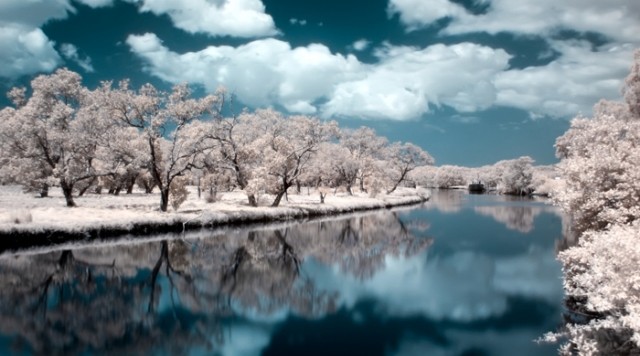}&
		\includegraphics[width=.24\linewidth]{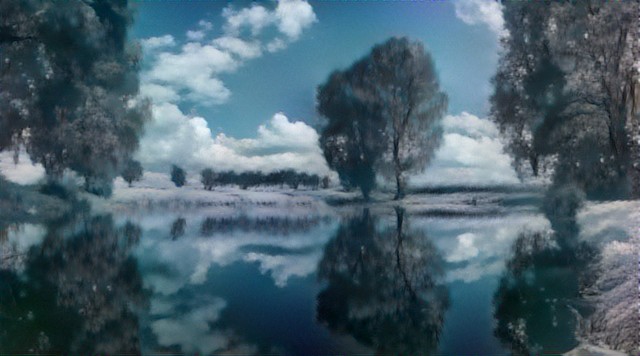}&
		\includegraphics[width=.24\linewidth]{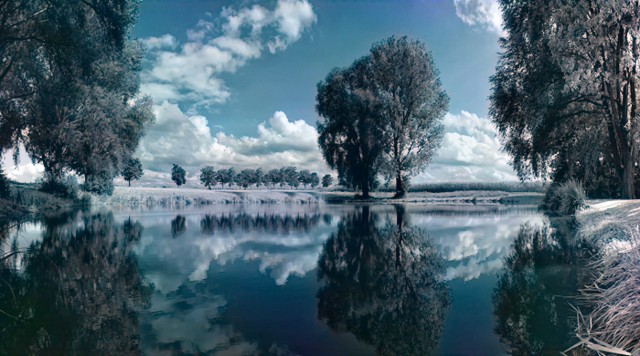}\\
		\includegraphics[width=.24\linewidth]{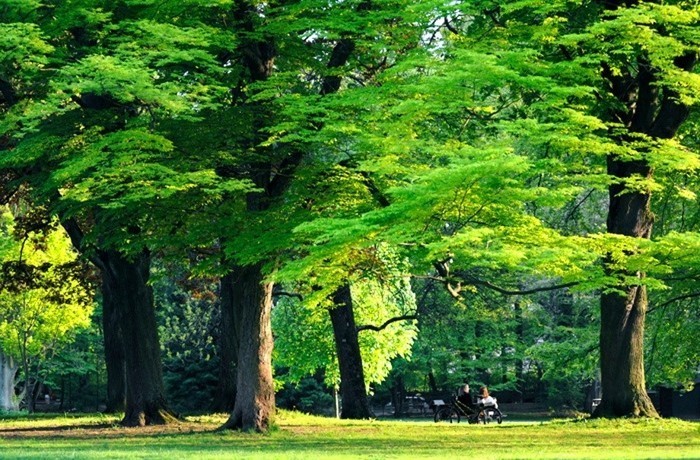}&
		\includegraphics[width=.24\linewidth]{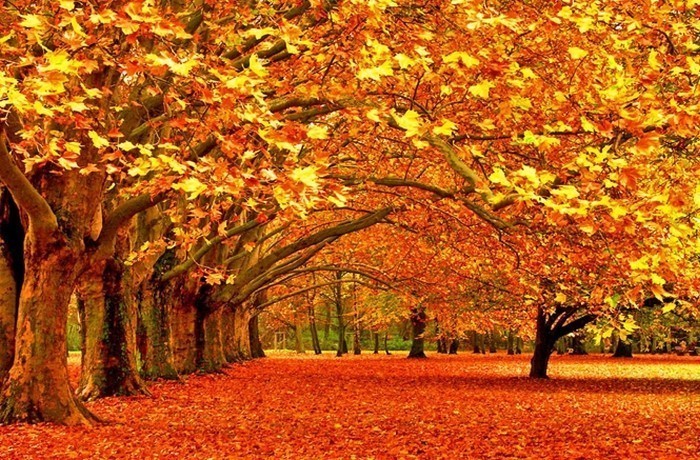}&
		\includegraphics[width=.24\linewidth]{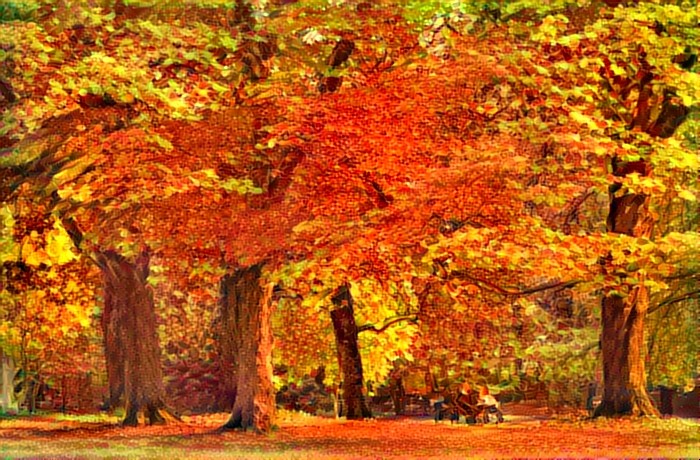}&
		\includegraphics[width=.24\linewidth]{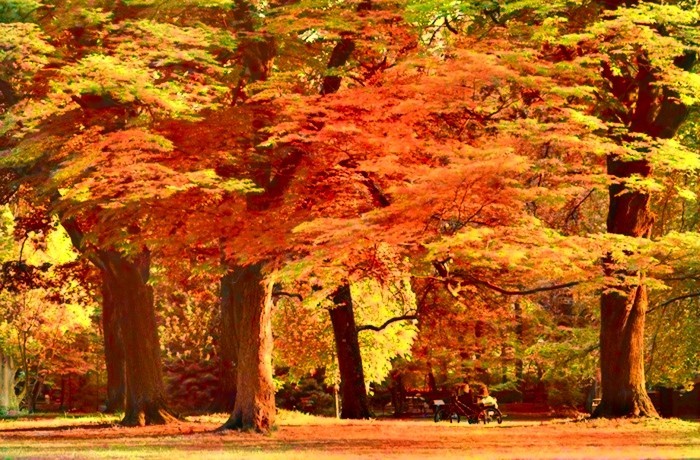}\\
        (a) Input image & (b) Style image & (c) Stylized image & (d) Our result
        \end{tabular}
		\caption{Classic style-transfer methods take an input image (a) and a reference style image (b) and produce a stylized image (c), typically showing texture artifacts and missing details that make it look like a painting. Our method processes the stylized image (c) and makes it photo-realistic (d). The identity of the original image is preserved while the desired style is reliably transferred. The styled images were produced by StyleSwap \cite{chen2016fast} (top) and NeuralStyle \cite{Gatys_2016_CVPR} (bottom). Best seen enlarged on a full screen.}
		\label{fig:teaser}
\end{figure}

Practitioners often use complex manipulations when editing a photo. They combine multiple effects such as exposure, hue change, saturation adjustment, and filtering, to produce a stylized image. The overall manipulation is complex and hard to reproduce.
Style transfer methods aim to resolve this by allowing an image editor (or a novice user) a simple way to control the style of an image, by automatically transferring the style of a reference image onto another image. 

Recently, several solutions have been proposed for style transfer~\cite{Gatys_2016_CVPR,li2016combining,chen2016fast,johnson2016perceptual,ulyanov2016instance}, producing stunning results.   
However, as impressive the resultant images are, their appearance is still non-photorealistic, painting-like, precluding style transfer methods from becoming a handy tool for photo editing. 
The challenges in image manipulation are hence dual: (i) achieving the desired artistic effect, and  (ii) producing a natural looking, photorealistic image. Current style transfer methods do pretty well on the former, but fail on the latter. 

%{\color{red}The first reason for the this limitation is the textures which transfused along the image, e.g. mosaic texture in the sky or blue transfers from Van Gogh "Starry Night" painting in a city scape content image. The second reason is colors which transfer in a non semantic manner, that is, we want the style color of a car to transfer to the content car and not tho the sidewalk and the road.}

Luan et al~\cite{luan2017deep} made a similar observation and proposed to resolve this via a two-stage optimization. The first step transfers the style using NeuralStyle of~\cite{Gatys_2016_CVPR}.
The second post-processing stage constrains the transformation to be locally affine in colorspace.
As stated in~\cite{luan2017deep}, this two-stage optimization works better than solving their objective directly, as it prevents the suppression of proper local color transfer due to the strong photorealism regularization.
They succeed in improving photorealism, motivating us to seek even better solutions.
 
Similarly to~\cite{luan2017deep} we prefer a two-stage approach, starting with style transfer and then post-processing to refine the photorealism. The post-processing stage we propose is a simple, yet effective. It takes a stylized image and manipulates it, making its appearance more photorealistic, using information from the original input image. Example results are presented in Figure~\ref{fig:teaser}.

The approach we propose makes three key contributions. First, we produce photorealistic and natural looking results by constraining the manipulation with the original image gradients. With respect to previous methods for style transfer, our output images are sharper, exhibit more fine details and fewer color artifacts. Second, in comparison to the outstanding work of Luan et al.~\cite{luan2017deep} (that also post-processes a stylized image), our approach is much faster, taking less than 2 seconds to run, in comparison to more than 2 minutes in~\cite{luan2017deep}. 
Finally, our method is very simple and can be applied at post-processing to any stylized image, regardless of the style transfer algorithm used to generate it.

%and have the potential to run on real time by using fast Screen Poisson solver based convolutional filter \cite{farbman2011convolution} or Fourier transform \cite{bhat2008fourier}. 
%Third, our approach is fully automatic and does not require user segmentation or parameter tuning.

The algorithm we propose is based on the Screened Poisson Equation (SPE) originally introduced by Bhat et al.~\cite{bhat2008fourier} for image filtering. Later on, Darabi et al.~\cite{darabi2012image}, used it for combining inconsistent images and for image completion. In their optimization, the colors and the gradients are generated separately and then combined using SPE. In~\cite{morel2014screened} SPE is used for image contrast enhancement. Mechrez et al.~\cite{mechrez2016saliency} suggested to use similar ideas in order to enforce photorealism in saliency-based image manipulation. Our method follows this line of work as we use the SPE to edit and manipulate images in a photorealistic manner. 
%Differently from those methods, we use the original gradients of the image in the SPE solver and by that gain strong realism. Further more use the SPE along all three color channels in Lab color space.

%We suggest solving this limitation by using the semantic-style-transfer of Chen et al.~\cite{chen2016fast} which based on Deep space nearest neighbor search and thus reduce the second foreign color problem. To overcome the texture distortions we suggest using SPE using the original image gradients. Our code will be publicly available.

%The rest of the paper is organized as follows. We start by surveying related work in Section~\ref{sec:related}. Next, in Section~\ref{sec:problem}, we provide a mathematical formulation for SPE and describe our method. Our empirical evaluations and results are presented in Section~\ref{sec:evaluation} and conclusions are drawn in Section~\ref{sec:conc}.

\vspace*{-0.3cm}
\section{Related Work}
\label{sec:related}

\paragraph{Artistic Style Transfer.} Style transfer between one image to another is an active field of research and many have tried to solve it, e.g. Hertzmann et al. \cite{hertzmann2001image}. Most recent approaches are based on CNNs, differing in the optimization method and the loss function~\cite{Gatys_2016_CVPR,li2016combining,chen2016fast,johnson2016perceptual,ulyanov2016instance}. Approaches which do not rely on CNNs have also been proposed~\cite{liang2001real,elad2017style,frigo2016split}, however, it seems like the use of deep feature space in order to transfer image properties gives a notable gain in this task.

Gatys et al.~\cite{Gatys_2016_CVPR} transfer style by formulating an optimization problem with two loss terms: style textures statistics and content reconstruction. The optimization is done using back-propagation and a gradient based solver. They allow arbitrary style images and produce stunning painterly results, but at a high computational cost.
Since then several methods with lower computation time have been proposed~\cite{dumoulin2017learned,ulyanov2016texture,ulyanov2016instance,johnson2016perceptual}. The speedup was obtained by training a feed-forward style network using a similar loss function to that of~\cite{Gatys_2016_CVPR}. The main drawback of these latter methods is that they need to be re-trained for each new style. 

In methods such as~\cite{Gatys_2016_CVPR} and~\cite{johnson2016perceptual} no semantic constraints exist, resulting in troubling artifacts, e.g., texture from a building or a road could be transferred to the sky and vise versa. 
Chen et al.~\cite{chen2016fast} and Huang et al.~\cite{huang2017arbitrary} suggested methods which are based on transferring statistical properties from the style image to the content image in the deep space and then inverting the features using efficient optimization or through a pre-trained decoder. 
They find for each neural patch its nearest neighbor -- this process implicitly enforces semantic information. 
Similarly, Li et al.~\cite{li2016combining} combine Markov Random Field (MRF) and CNN (CNNMRF) in the output synthesis process. CNNMRF results are much more realistic than those of other style transfer methods, however, the stylization effect is not as strong.

Common to all of these methods is that the stylized images are non-photorealistic and have a painting-like appearance. 

\paragraph{Realistic Style Transfer.} Recently Luan et al.~\cite{luan2017deep} proposed a deep-learning method to transfer photographic style to a content image in a realistic manner. Painting-like artifacts are overcome by: (i) Semantic segmentation is used to make sure the style is being transferred only between regions with similar labels. (ii) The transformation from the input to the output is constrained to be locally affine in color-space. The second component can also be seen as a post-processing that is based on the \textit{Matting Laplacian} (ML) regularization term of~\cite{levin2008closed}. This method produces realistic output images and the style is transferred faithfully. On the down side, the computation is rather slow, taking over 2 minutes per image.

Other approaches to realistic style transfer were limited to specific problem and style, such as faces and time-of-day in city scape images \cite{shih2013data,shih2014style}.

\paragraph{Color Transfer.}
Our method is also related to methods for global color transfer and histogram matching. For instance, Wu et al.~\cite{wu2013content} transfer the color patterns between images using high-level scene understanding. In \cite{pitie2005n,reinhard2001color,pitie2007automated} a more statistic approach is taken for matching the color mean, standard deviation or 1D histograms. These methods lack semantic information. Colors are often wrongly transferred and textures are not generated properly.

Last, our method is also related to other photo-realistic manipulations such as \cite{sunkavalli2010multi,laffont2014transient,gardner2015deep,bae2006two}. Our work differs from these methods in its generality -- it can work with any style image and any style-transfer prior.

\vspace*{-0.3cm}
\section{Method}
\label{sec:problem}
The approach we propose consists of two stages. Given a content image $C$ and a style image $S$ we first produce a stylized version denoted by $C_S$. We then post-process $C_S$ with a solver based on the Screened Poisson Equation (SPE), resulting in a photorealistic, stylized, output image $O$. 
We next describe the methods we have used for generating the stylized image and then proceed to describe our SPE based post-processing.

%========================================================
\vspace*{-0.4cm}
\subsection{Stage 1: Deep Style Transfer}
For completeness, we briefly describe three style transfer methods that could be used as a first stage of our overall framework. Each of these three approaches has its own pros and cons, therefore, we experiment with all three.

\vspace*{-0.2cm}
\paragraph{NeuralStyle+Segmentation} 
Gatys et al.~\cite{Gatys_2016_CVPR} achieved groundbreaking results in painterly style transfer. Their method, called NeuralStyle, employs the feature maps of discriminatively trained deep convolutional neural networks such as VGG-19~\cite{simonyan2014very}. 
As impressive as its results are, NeuralStyle suffers from two main drawbacks: (i) its results are not photorealistic, and (ii) it lacks semantics, e.g., it could, for example, transfer the style of greenery to a vehicle. 
Luan et al.~\cite{luan2017deep} resolve the latter by integrating NeuralStyle with semantic segmentation to prevent the transfer of color and textures from semantically different regions. 
Luan et al.~\cite{luan2017deep} also attempt to improve the photorealism of the stylized images via a post-processing step based on the Matting Laplacian of~\cite{levin2008closed}. 
To compare to~\cite{luan2017deep} we tested using the same NeuralStyle+Segmentation (NS+segment) algorithm to generate a stylized image, and replaced their post-processing stage with the one we propose below. 

\vspace*{-0.4cm}
\paragraph{StyleSwap}
An alternative approach to incorporate semantics is to match each input neural patch with the most similar patch in the style image to minimize the chances of an inaccurate transfer. This strategy is essentially the one employed by StyleSwap~\cite{chen2016fast}, an optimization based on local matching that combines the content structure and style textures in a single layer of a pre-trained network. It consists of three steps
\textit{(i) Encoding:} Using a pre-trained CNN, such as VGG-19~\cite{simonyan2014very}, as the encoder $\mathcal{E}$, the content image $C$ and style image $S$ are encoded in deep feature space.
\textit{(ii) Swapping:} Each neural patch of $C$ is replaced with its Nearest Neighbor (NN) neural patch of $S$ under the cosine distance yielding $NN(C|S)$, the deep representation after the patch swapping. 
\textit{(iii) Decoding:} The decoding stage $\mathcal{D}$ inverts the new feature representation back to image space using either an optimization process or a pre-trained decoder network~\cite{chen2016fast}. The loss for this inversion process conserves the deep representation of the image: $\mathcal{L}=||\mathcal{E}(\mathcal{D}(NN(C|S))) - NN(C|S)||^2$.
A major advantage of StyleSwap over NS+segment~\cite{luan2017deep} is a much lower computation time.  However, a major limitation of StyleSwap over NS+segment~\cite{luan2017deep} is its ability in transferring the style faithfully.

\vspace*{-0.4cm}
\paragraph{CNNMRF}
Similar in spirit to StyleSwap is the CNNMRF of~\cite{li2016combining}, where a patch search, borrowed from MRF, is used to regularize the synthesis process instead of the Gram Matrix statistics used in~\cite{Gatys_2016_CVPR}. CNNMRF yields more photorealistic images, compared to NeuralStyle, however, it is prone to artifacts due to local mismatch between the nearest neighbor patches or global mismatch between the content and style images.

%========================================================
\subsection{Stage 2: Photorealism by the Screened Poisson Equation} 
\begin{figure}
\centering
\tabcolsep=0.13cm
\scriptsize
\begin{tabular}{cc}
				\bmvaHangBox{\includegraphics[width=.35\linewidth]{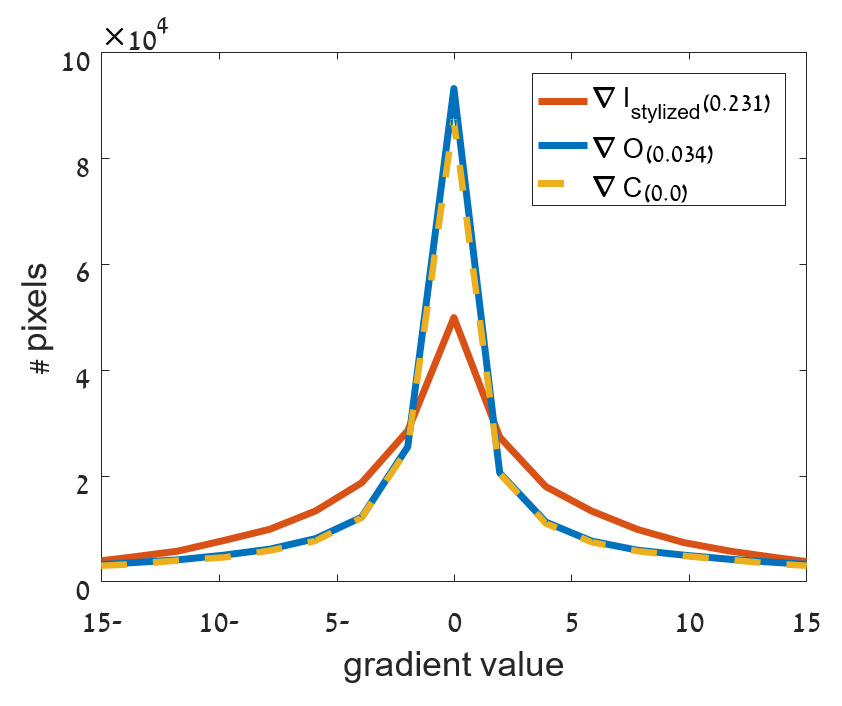}}&
				\bmvaHangBox{
				\tabcolsep=0.13cm
        \begin{tabular}{ccc} 
        \includegraphics[width=.15\linewidth]{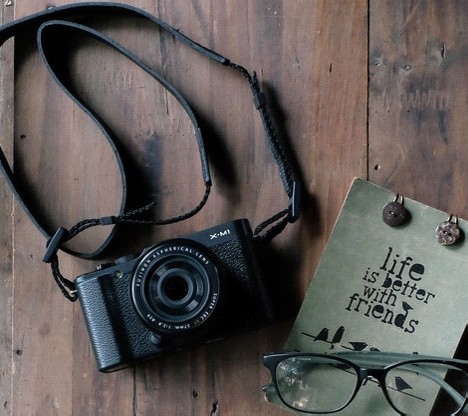}&
				\includegraphics[width=.15\linewidth]{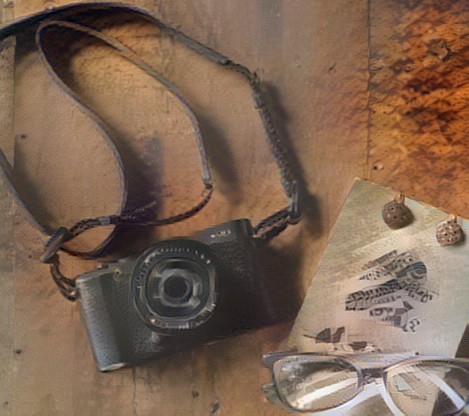}&
				\includegraphics[width=.15\linewidth]{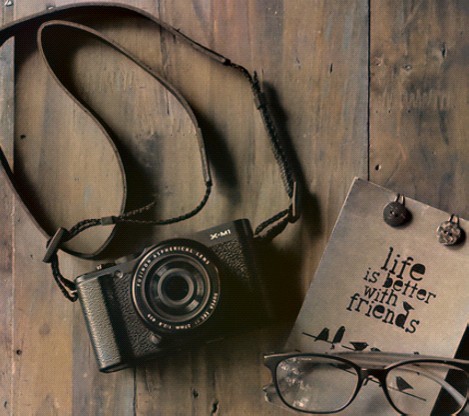}\\
				\includegraphics[width=.15\linewidth]{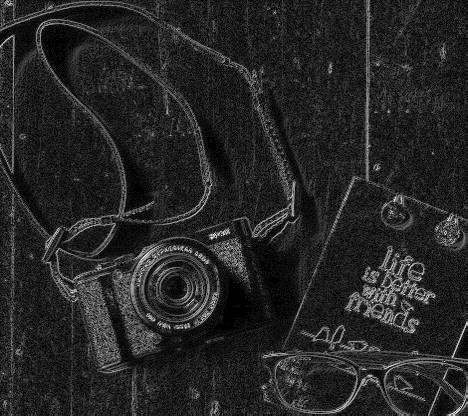}&
				\includegraphics[width=.15\linewidth]{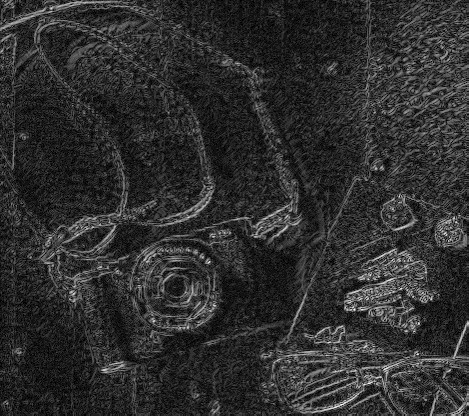}&
				\includegraphics[width=.15\linewidth]{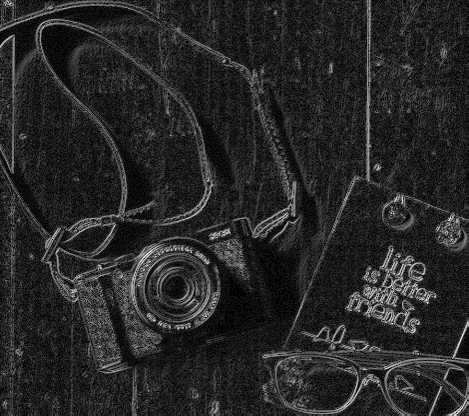}\\
        (a) Input image & (b) Stylized image & (c) Our result
        \end{tabular}}
\end{tabular}
		\caption{\textbf{Image Gradients} \textit{Left:} The distribution of gradients of natural images is known to be Laplacian (yellow curve). The distribution of gradients of a typical stylized image (red curve) is different, that can explain their non-photorealistic look. Images generated via our approach, on the other hand, share the statistics of natural images (blue curve). KL-Divergence distances with respect to the histogram corresponding to the input image are shown in the legend. \textit{Right:} An input image (a), it's stylized version using NeuralStyle (b), and our result (c), and their corresponding gradients below. NeuralStyle result lacks fine details, on one hand, while showing gradients in areas that should have none, on the other hand.}
		\label{fig:grad}
\end{figure}

When stylizing images with real photos as the style image, the results images generated in Stage 1 typically pertain severe visual artifacts that render them non-photorealistic. There are three common types of artifacts, as illustrated in Figure~\ref{fig:grad}.
The first is unnatural textures that appear in image regions that should be homogeneous. 
The second is distortion of image structures, e.g., wiggly lines instead of straight ones.
Third, fine details are often missing resulting in cartoon-like appearance.
The goal of Stage 2 is to get rid these artifacts. 

Common to the three types of artifacts is that all imply wrong image gradients. 
Homogeneous regions should have low gradients, structures such as lines should correspond to gradients in specific directions, and overall photorealistic appearance requires certain gradient domain statistics \cite{weiss2001deriving}. 
Figure~\ref{fig:grad} illustrates these artifacts on an example image.

To correct the gradients, we assert that the gradient field of the original content image comprises a good prior for correcting the stylized image. 
We would like to correct the gradients of the stylized image, by making them more similar to those of the input image, while at the same time retaining the transferred style colors from Stage 1. 
This suggests the usage of an objective with two terms. A fidelity term that requires similarity to the stylized image and its \textit{style properties}, and a gradient term that requires similarity of the gradients to those of the input image and its \textit{realistic properties}.
As we show next, an effective way to combine these two components is via the \textit{Screened Poisson Equation}\cite{bhat2008fourier}.

%Originally SPE has been suggested to solve various classic image processing tasks in the gradient domain, such as blending \cite{darabi2012image}, manipulating \cite{bhat2008fourier,mechrez2016saliency} and contrast enhancement \cite{morel2014screened}. It is designed to integrate two terms: a fidelity term and a gradient term. In our case, the integration between the fidelity and gradient terms is equivalent to integrating the original image and with its \textit{realistic properties} and the stylized image with its \textit{style properties}.

We begin with the gradient term. Given the gradient field $\nabla C(x,y)$ of the content image, we would like to compute the function $O(x,y)$ which satisfies the following term:
\begin{equation}
\int_{\Omega} ||\nabla O - \nabla C(x,y) ||^2dxdy
\label{eq:grad_term}
\end{equation}
%We can derive from this the Euler-Lagrange equation:
%\begin{equation}
%\nabla ^2O=\nabla \cdot g
%\end{equation}
To integrate also the fidelity term into the objective, that is, to require that $O(x,y)$ is as close as possible to the stylized image $C_S(x,y)$ we modify the objective function in~\eqref{eq:grad_term}:
\begin{equation}
\mathcal{L} =  \int_{\Omega} ||O-C_S||^2 +  \lambda \cdot ||\nabla O - \nabla C(x,y) ||^2dxdy,
\label{eq:grad_data_term}
\end{equation}
where $\lambda$ is a constant that controls the relative weight between the two terms. Optimizing this objective leads to the Screened Poisson Equation:
\begin{equation}
O- \lambda \nabla ^2 O = C_S -  \lambda \nabla ^2 C(x,y)
\label{eq:poisson}
\end{equation}
or equivalently:
\begin{equation}
 (O-C_S) - \lambda(O_{xx}-C_{xx}) - \lambda(O_{yy}-C_{xx}) = 0
\label{eq:poissonO}
\end{equation}
This objective defines a set of linear equations that can be solved using Least Squares, Fourier transform~\cite{bhat2008fourier} or convolving~\cite{farbman2011convolution}.

%========================================================
We work in $Lab$ color space and solve the SPE for each channel separately using  $\lambda=5$ for the $L$ channel and $\lambda=1$ for $a,b$ channels. Note, that when $\lambda = 0$ we get $O=C_S$ and when $\lambda$ tends to infinity, we get $O=C$.

\begin{figure}[t]
		\setlength{\tabcolsep}{.16em}
        \scriptsize
        \begin{tabular}{cc|cccc} 
        \includegraphics[width=.16\linewidth]{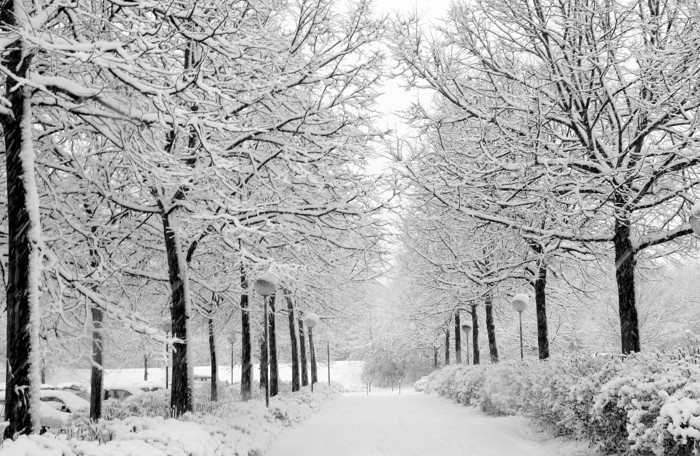}&
		\includegraphics[width=.16\linewidth]{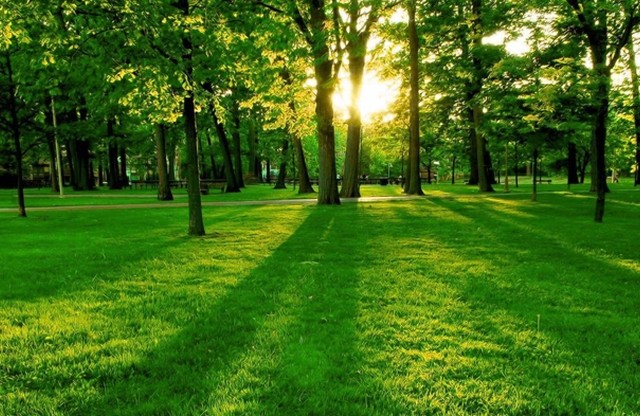}&
		\includegraphics[width=.16\linewidth]{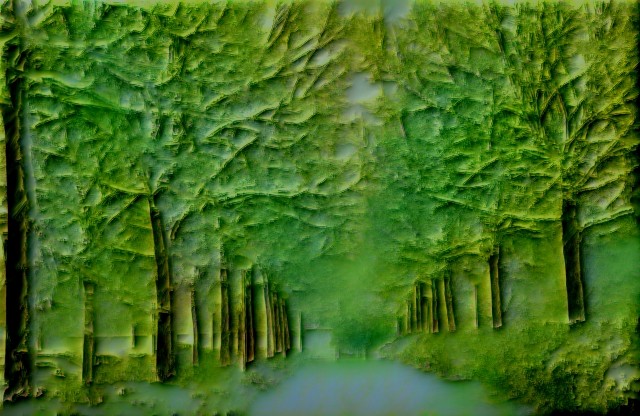}&
		\includegraphics[width=.16\linewidth]{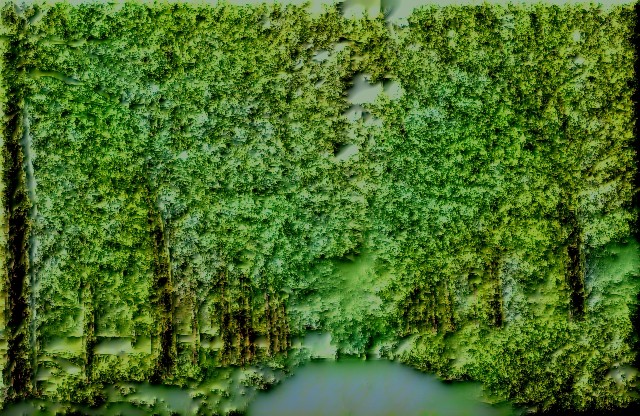}&
		\includegraphics[width=.16\linewidth]{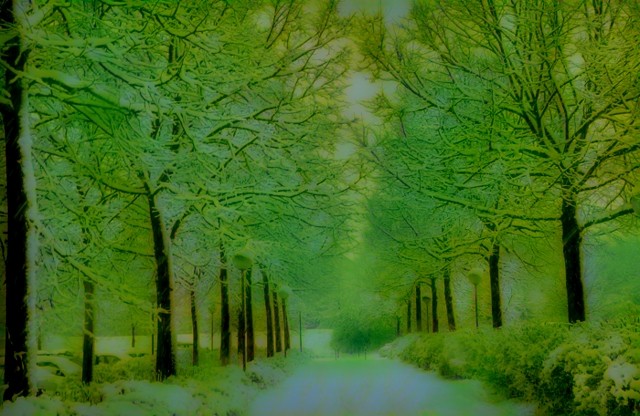}&
		\includegraphics[width=.16\linewidth]{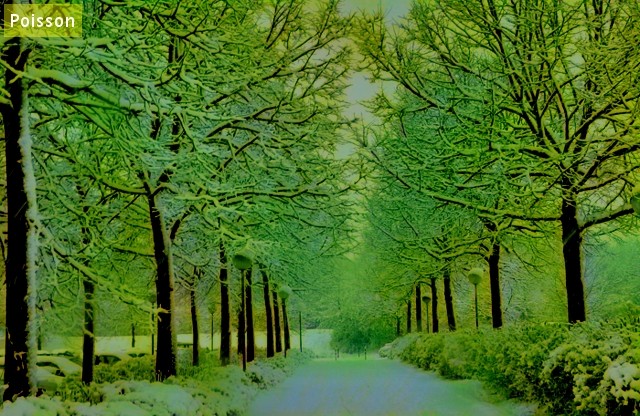}\\
        Input image& Style image& (a) $ SPE(abs(g))$& (b) $SPE(g^2)$ & (c) $SPE(HM(g))$ &(d) $SPE(g)$ \\
        \end{tabular}
		\caption{\textbf{The Gradient term.} Four alternative gradient terms were tested in the SPE solver: (a) absolute, (b) square, (c) histogram matching (HM) w.r.t the style image gradients and (d) original gradients. The latter was found to be the most successful.
        }
		\label{fig:poisson}
\end{figure}

%========================================================
\paragraph{Alternative gradients terms}
The gradient term we chose is not the sole option one could chose. 
We have additionally tested three alternative gradients terms (we mark $g=\nabla C(x,y)$ for clarity): (i) using absolute values $|| \mbox{\ } |\nabla O| - |g|  \mbox{\ } ||^2$, (ii) using squared gradients $|| \nabla O^2 - g^2 ||^2$, and, (iii) matching the histograms of gradients $|| h(\nabla O) - h(g) ||^2$.
All were found inferior. 
Figure~\ref{fig:poisson} shows an example comparison of these alternatives which illustrates that using the original gradients yields the best results. 
\vspace{-0.2cm}

%========================================================
%{\color{red}\paragraph{Dual Problem to $l_1$-Regularized Least Squares}
%We observed that in many cases in order to achieve faithful transfer of the style a drastic contrast change is needed.  For example in Figure~\ref{fig:o} bottom row, a large change in the contrast between the buildings and sky is noticed. Commonly, the gradient term in eq.~\ref{eq:grad_data_term} is taken as $L_2$ regularization. In order to allow more flexible contrast changes we replace it with the $L_1$ norm of the gradient term. The $L_1$ norm is more forgivable for large changes in some regions and thus avoids artifacts in the border between neighboring regions with high contrast. 
%In order to solve the new $L_1$ regularization Least-squares problem we convert it to the equivalent dual problem, were the regularization term is the gradient term. This conversion is described in detail in Boyd and Vandenberghe~\cite{boyd2004convex} and more specifically in \cite{schmidt2008two}.}

%========================================================
\subsection{RealismNet}
The SPE based post-processing proposed above successfully turns a stylized image into a photorealistic one, as we show via extensive experiments in Section~\ref{sec:evaluation}.
Encouraged by these results we have further attempted to train a deep network that would apply the same effect. 
The input to the net would be the stylized image $C_S$ and its output would be a photorealistic, and stylized, image $O$. 

We collected 2500 style-content image pairs (total of 5000 images) from Flicker using the following keywords: \textit{sea, park, fields, houses, city}, where the vast majority of images were taken outdoors. We applied StyleSwap followed by SPE to all image pairs. These were then used to train an image-to-image \emph{RealismNet} using Conditional Generative Adversarial Network (cGAN)~\cite{isola2016image}.
We used the ``U-Net'' architecture of~\cite{ronneberger2015u}, an encoder-decoder with skip connections between mirrored layers in the encoder and decoder stacks.
Our RealismNet succeeds in improving photorealism (results are available in the supplementary), however, it is not as effective as SPE. Hence, we currently recommend using SPE. 
One potential advantage of such a method would be that it does \emph{not} require constraining to the original image gradients. Hence, its utility could be broader also for other applications where currently the output images are not photorealistic. To add to this, an end-to-end solution is faster than the optimization based solver.

Another alternative we have explored is to use an end-to-end optimization (differ from a pretrained network) that will combine the stylization optimization with a photorealistic constraint.
This was done by adding a gradient-based loss term to the optimization of~\cite{chen2016fast} and~\cite{Gatys_2016_CVPR}.
Unfortunately, we found it very difficult to balance between the gradient term and the style term. When the weight of the gradient term is too large the style was not transferred to the content image. Conversely, when the gradient term weight was too small the results were not photorealistic. 
Our observations match those of~\cite{luan2017deep} (supplementary) who report that applying their Matting Laplacian post-processing as part of the optimization of NeuralStyle was not successful. We conclude that an end-to-end solution to photorealistic deep style transfer remains an open question for future research.

\vspace*{-0.3cm}
\section{Empirical Evaluation}
\label{sec:evaluation}

\begin{figure}
		\centering
        \setlength{\tabcolsep}{.15em}
        \scriptsize
        \begin{tabular}{cc|ccc} 
        \includegraphics[width=.19\linewidth]{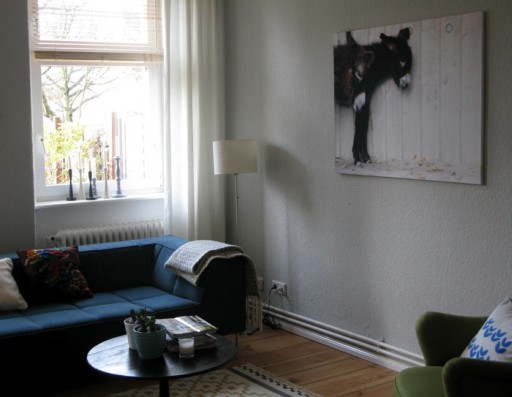}&
		\includegraphics[width=.19\linewidth]{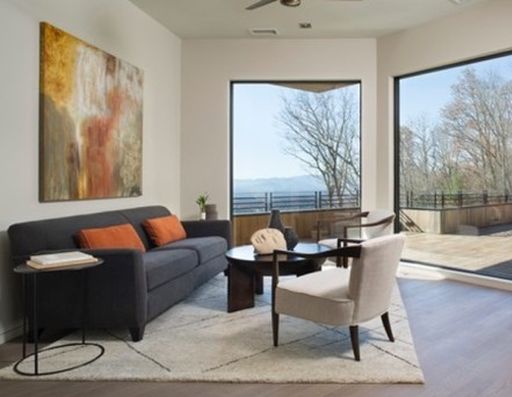}&
        \includegraphics[width=.19\linewidth]{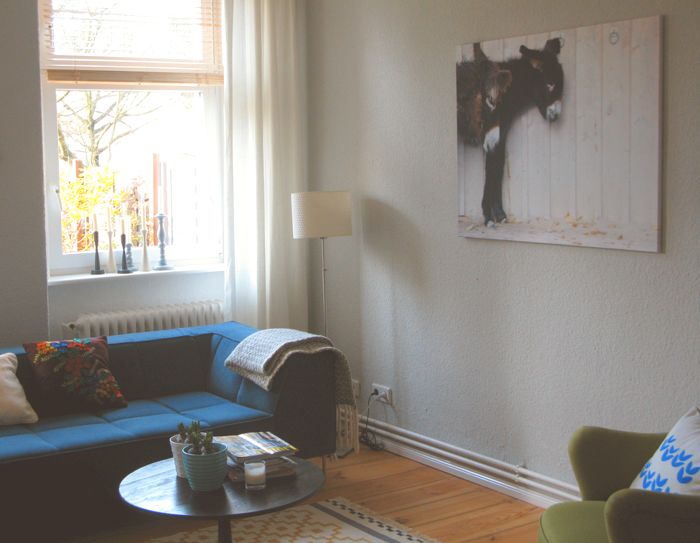}&
        \includegraphics[width=.19\linewidth]{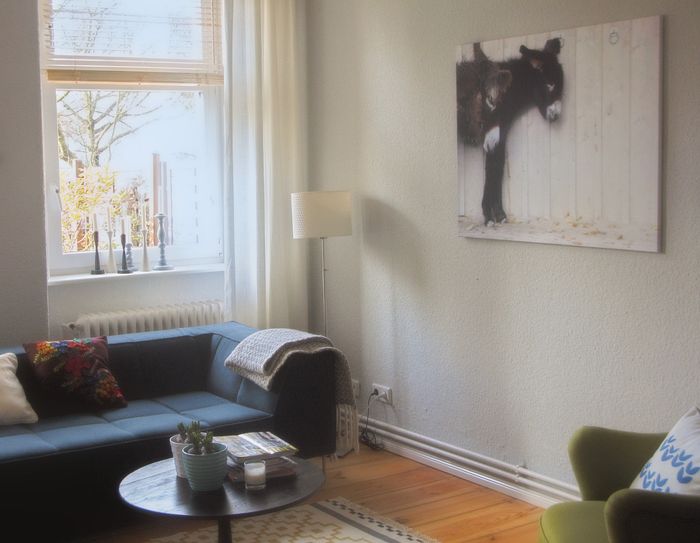}&
		\includegraphics[width=.19\linewidth]{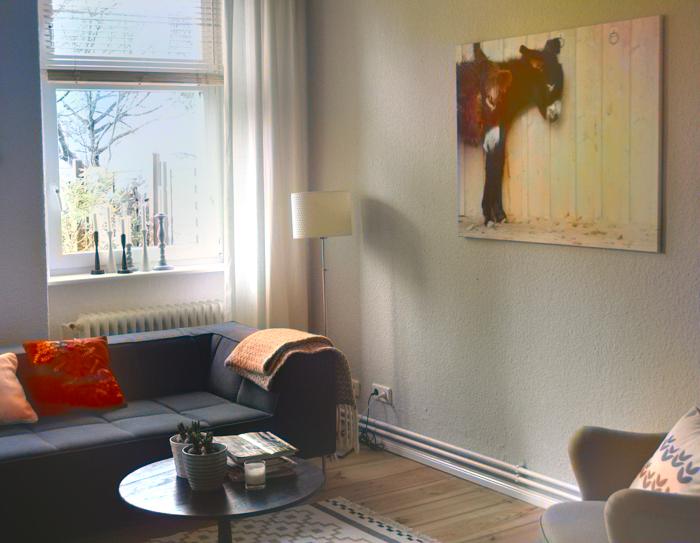}\\
        Input image& Style image& Pitie et al. 07~\cite{pitie2007automated}& Pitie et al. 05 ~\cite{pitie2005n}& Ours result
        \end{tabular}
                \begin{tabular}{cc|ccc} 
		\includegraphics[width=.19\linewidth]{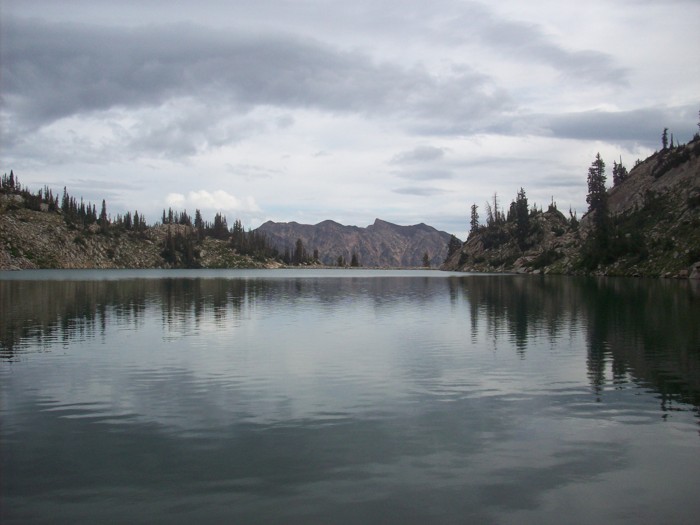}&
		\includegraphics[width=.19\linewidth]{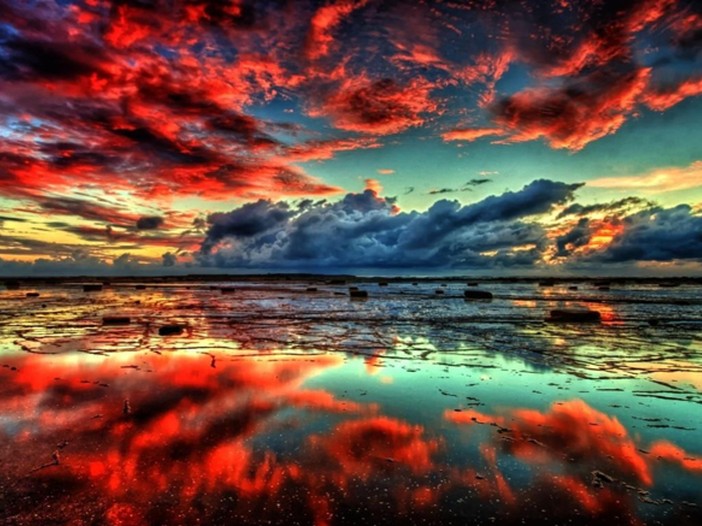}&
		\includegraphics[width=.19\linewidth]{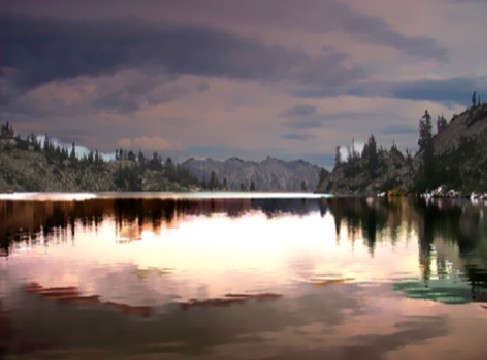}&
        \includegraphics[width=.19\linewidth]{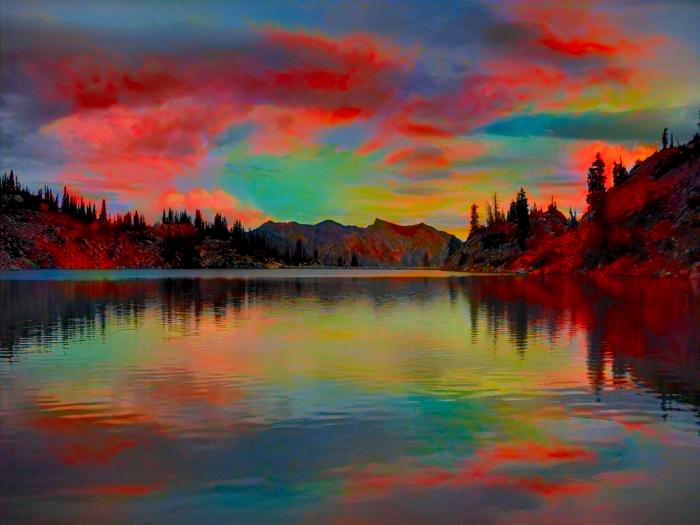}&     
        \\
		\includegraphics[width=.19\linewidth]{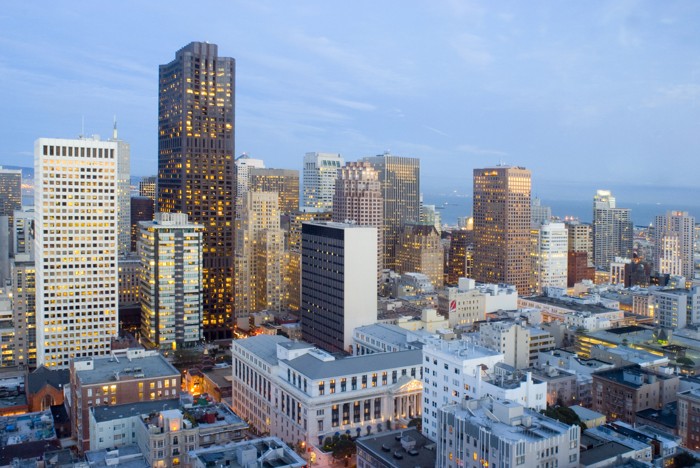}&
		\includegraphics[width=.19\linewidth]{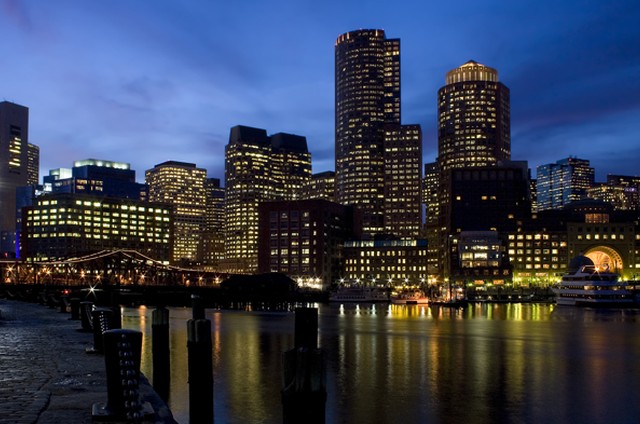}&
        \includegraphics[width=.19\linewidth]{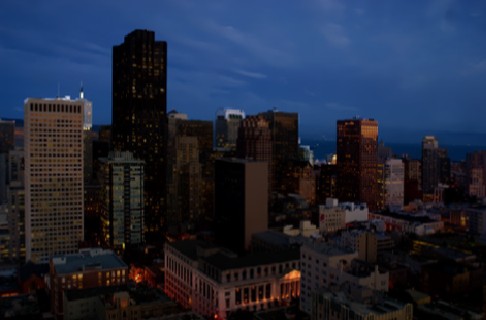}&
        \includegraphics[width=.19\linewidth]{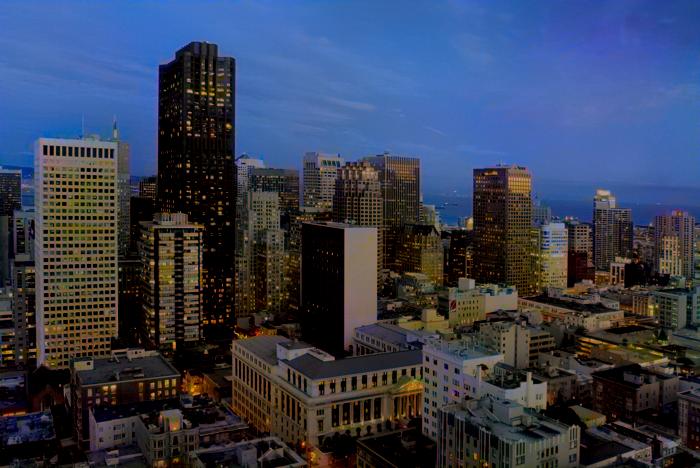}&
        \\
        Input image& Style image& Wu et al.~\cite{wu2013content}& Ours result & \\
        \end{tabular}
		\caption{\textbf{Comparison with color transfer.} Color transfer results are photorealistic, however, those that ignore semantics~\cite{pitie2007automated,pitie2005n} could fail in transferring the style faithfully, e.g. the pillow and painting color and the room light.
        %also produce unnatural haze-like artifacts. 
        Content-aware methods~\cite{wu2013content} do not transfer full style properties, e.g., the building windows are not lit and the patterns on the lake were not transferred. In our results we use NeuralStyle+segmentation combined with SPE. 
        }
		\label{fig:compare}
\end{figure}

%==========================================
\vspace*{-0.1cm}
To evaluate photorealistic style transfer one must consider two properties of the manipulated image: (i) the style faithfulness w.r.t the reference style image, and, (ii) photorealism. We compare our algorithm to DPST~\cite{luan2017deep}, that also aim at photorealism, through these two properties. We also show that SPE can be combined with StyleSwap~\cite{chen2016fast}, CNNMRF~\cite{li2016combining} and NeuralStyle+segmentation~\cite{luan2017deep}. 
Our experiments used the data-set of~\cite{luan2017deep}. 

\begin{figure}
		\centering
		\setlength{\tabcolsep}{.15em}
        \small
        \begin{tabular}{cc|cc} 
		\includegraphics[width=.19\linewidth]{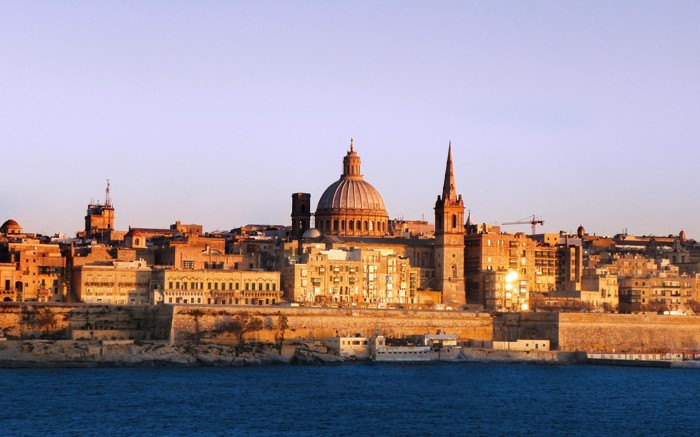}&
		\includegraphics[width=.19\linewidth]{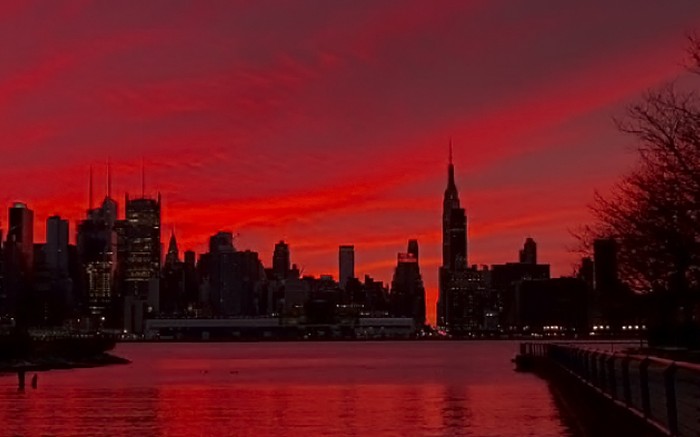}&
		\includegraphics[width=.19\linewidth]{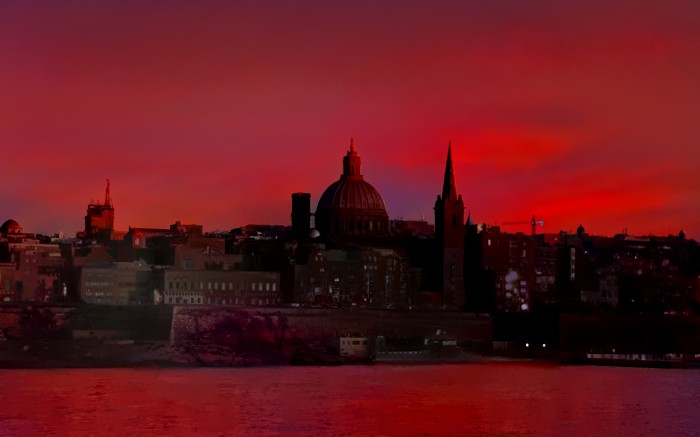}&
        \includegraphics[width=.19\linewidth]{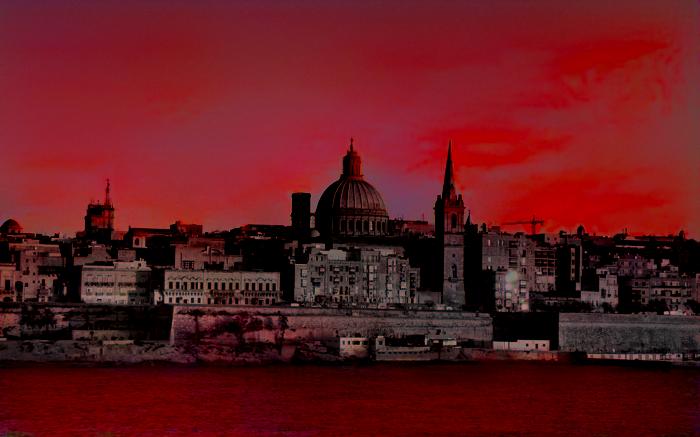}\\
        \includegraphics[width=.19\linewidth]{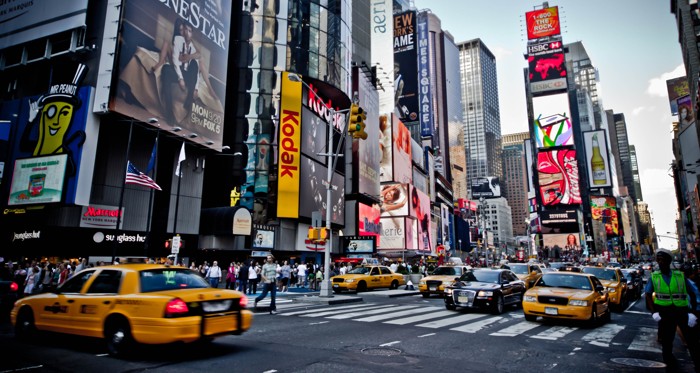}&
		\includegraphics[width=.19\linewidth]{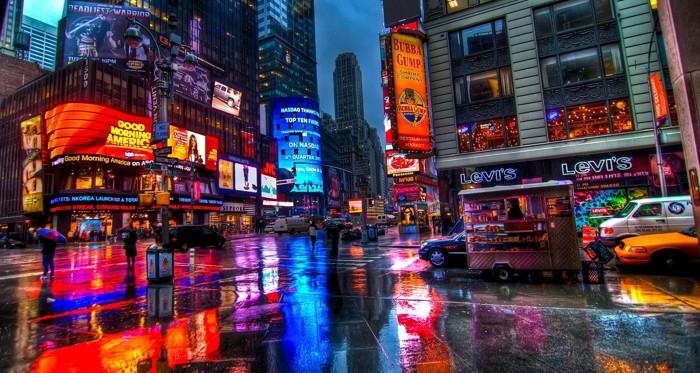}&
		\includegraphics[width=.19\linewidth]{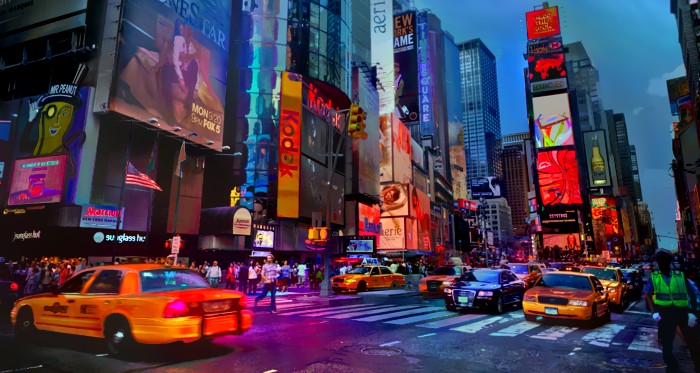}&
        \includegraphics[width=.19\linewidth]{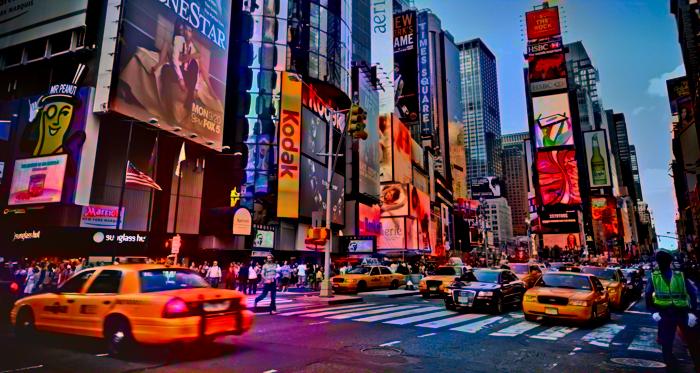}\\
        \includegraphics[width=.19\linewidth]{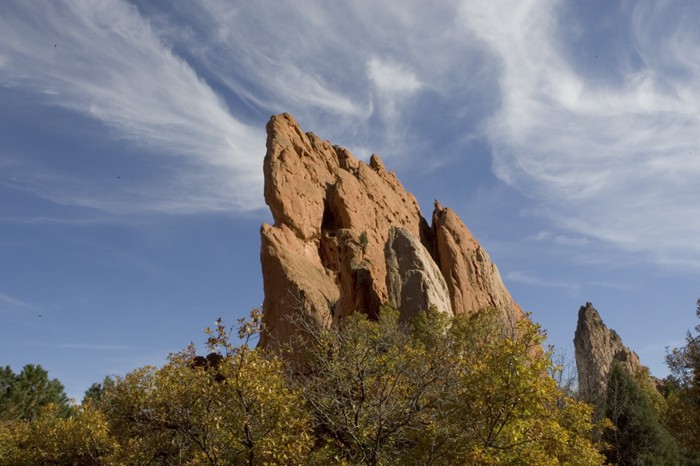}&
		\includegraphics[width=.19\linewidth]{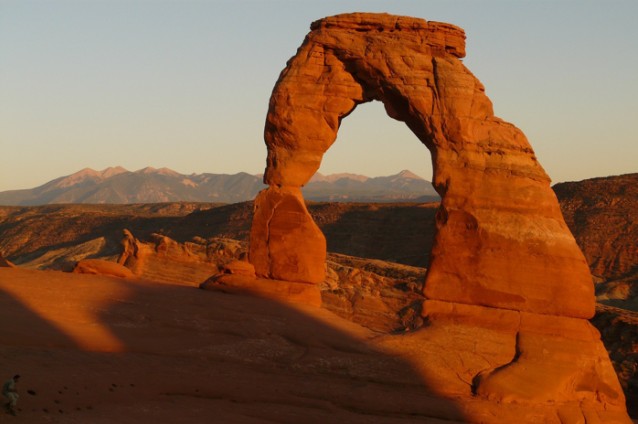}&
		\includegraphics[width=.19\linewidth]{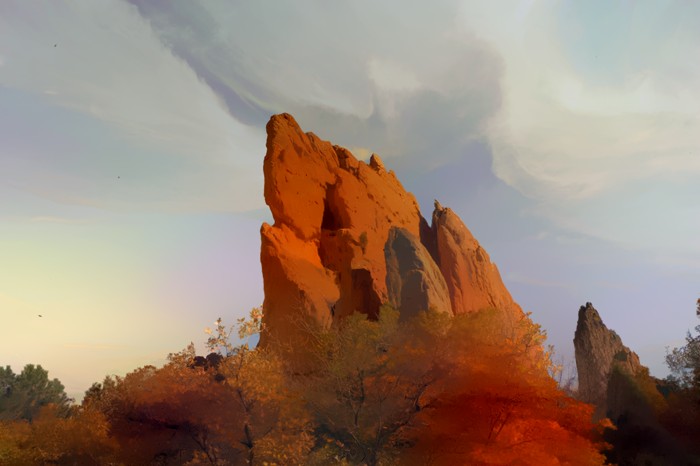}&
        \includegraphics[width=.19\linewidth]{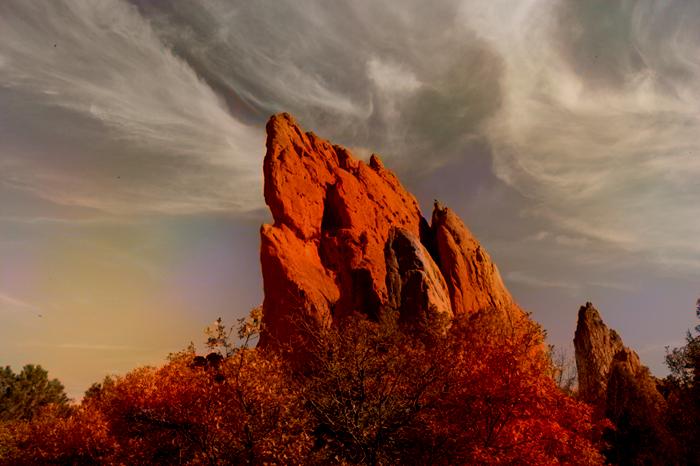}\\
        \includegraphics[width=.19\linewidth]{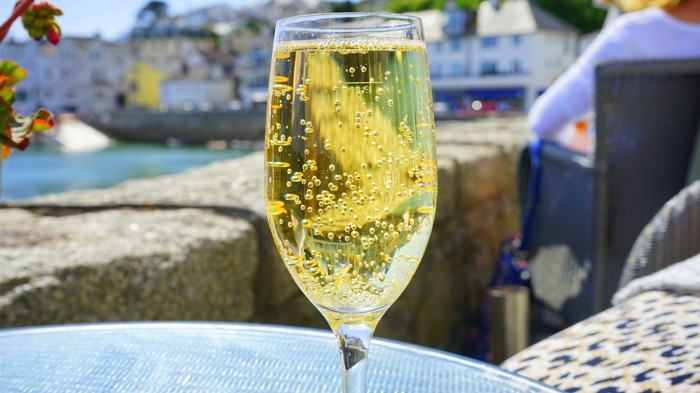}&
		\includegraphics[width=.19\linewidth]{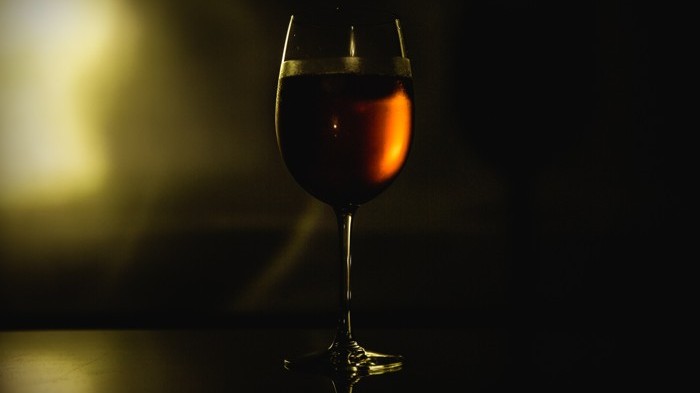}&
		\includegraphics[width=.19\linewidth]{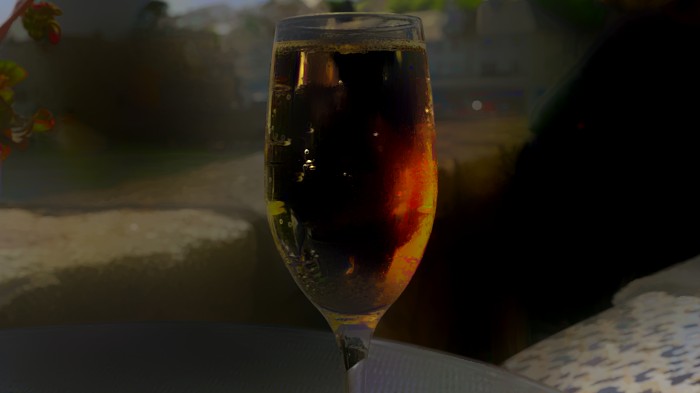}&
        \includegraphics[width=.19\linewidth]{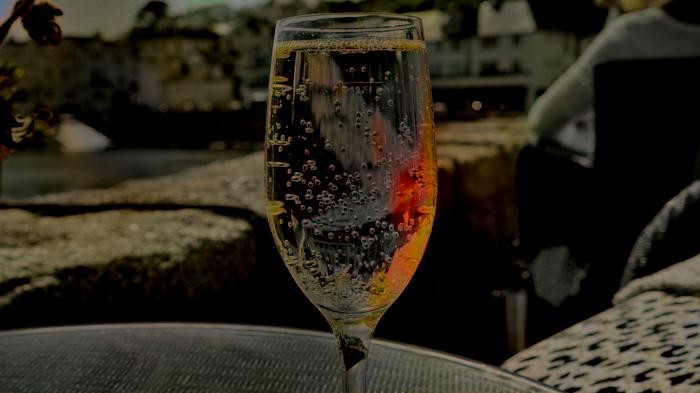}\\
        \includegraphics[width=.19\linewidth]{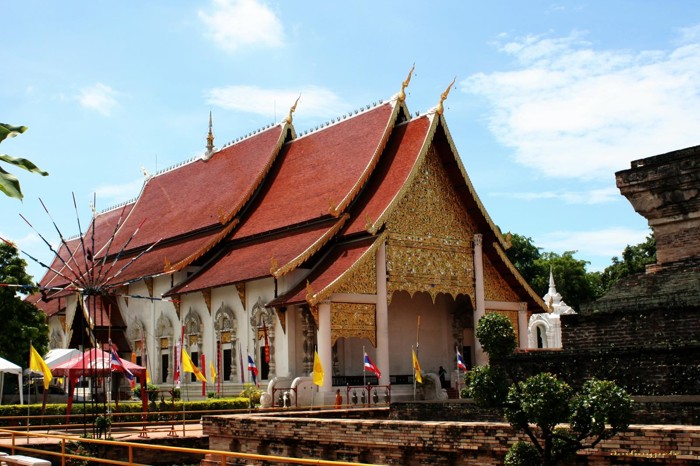}&
		\includegraphics[width=.19\linewidth]{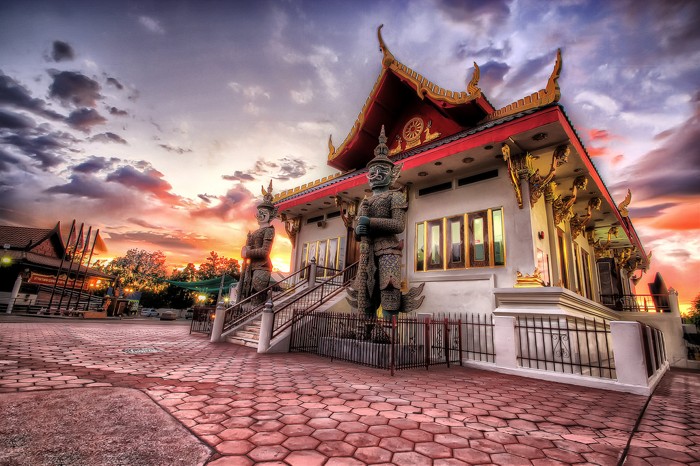}&
		\includegraphics[width=.19\linewidth]{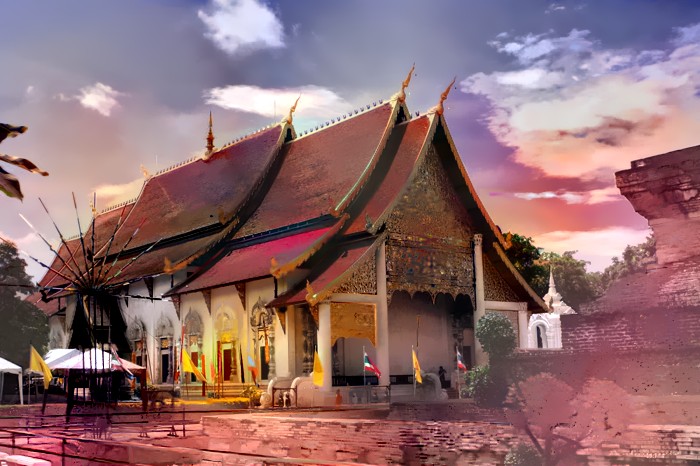}&
        \includegraphics[width=.19\linewidth]{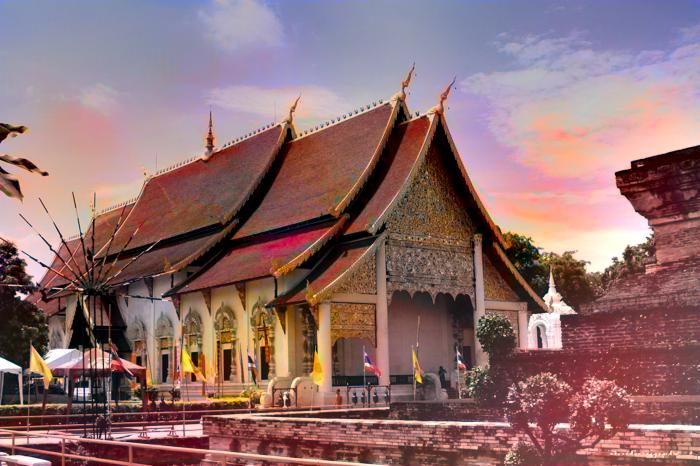}\\
        \includegraphics[width=.19\linewidth]{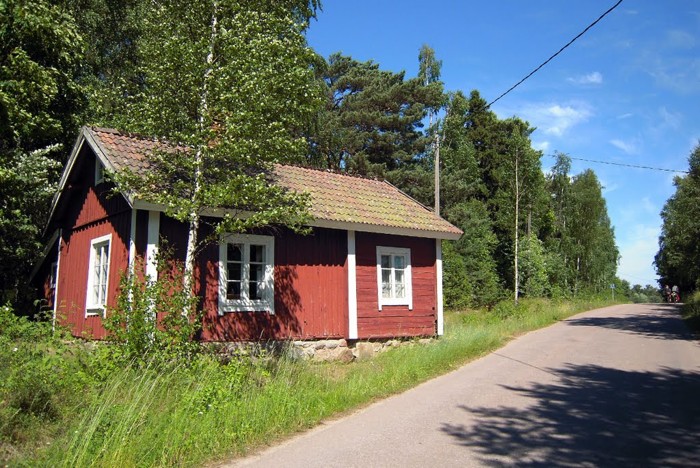}&
		\includegraphics[width=.19\linewidth]{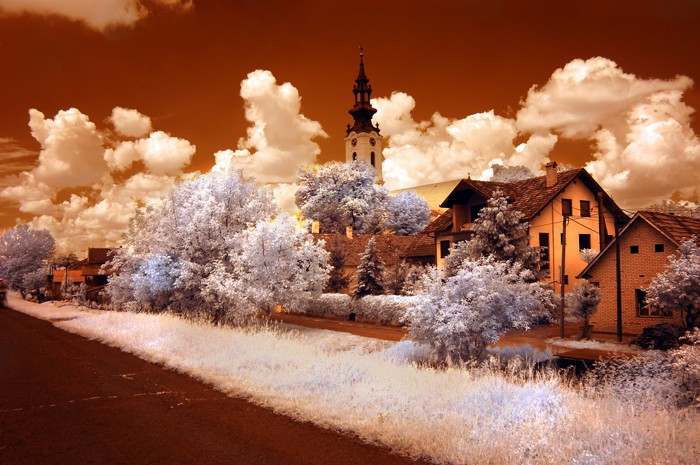}&
		\includegraphics[width=.19\linewidth]{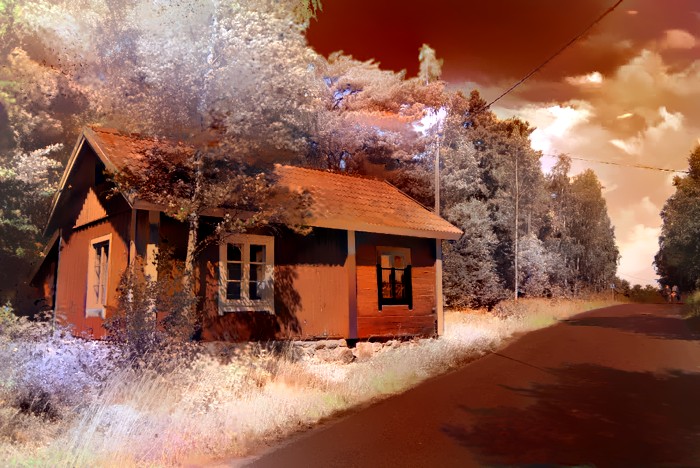}&
        \includegraphics[width=.19\linewidth]{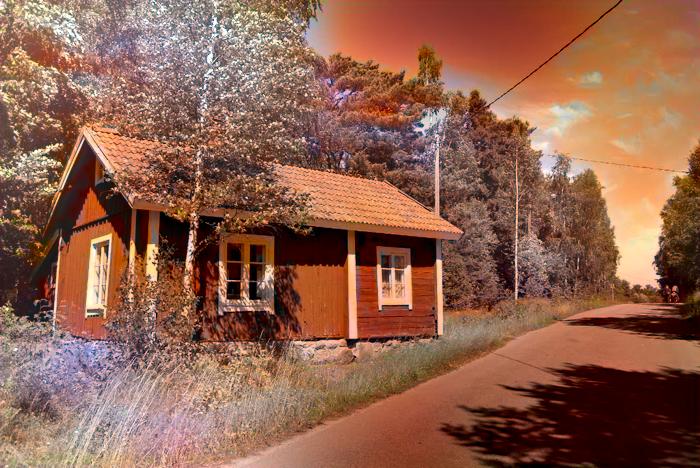}\\
        \includegraphics[width=.19\linewidth]{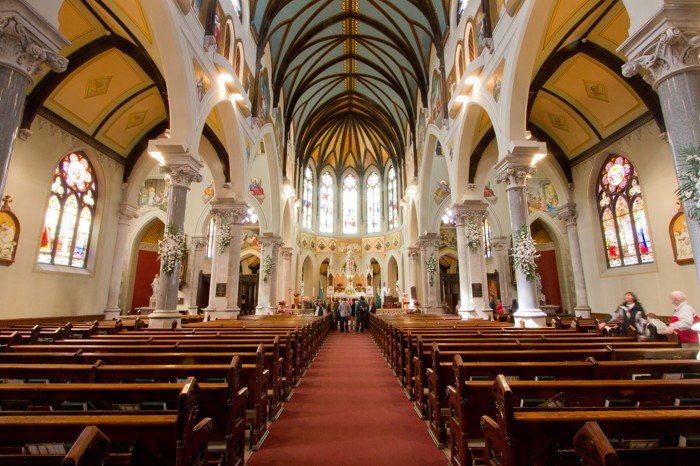}&
		\includegraphics[width=.19\linewidth]{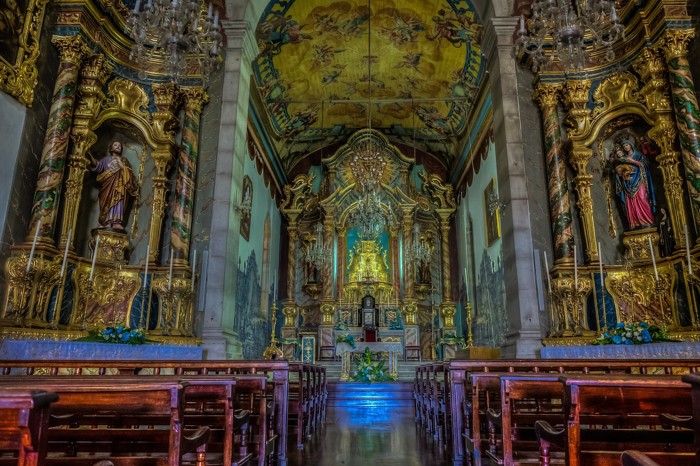}&
		\includegraphics[width=.19\linewidth]{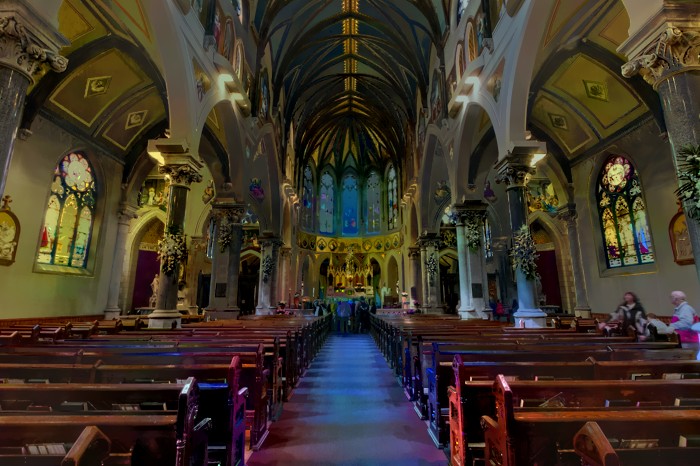}&
        \includegraphics[width=.19\linewidth]{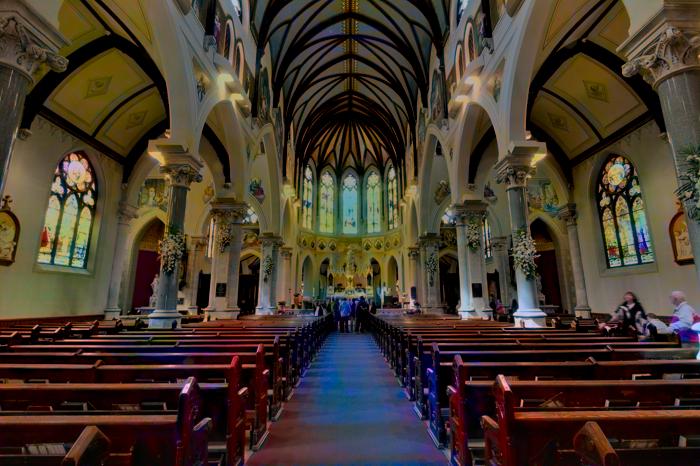}\\
        Input image & Style image & DPST~\cite{luan2017deep} & NS+segment+SPE %&StyleSwap+SPE
        \end{tabular}
\caption{\textbf{Qualitative assessment.} SPE preserves fine details and textures in the image and prevents smoothing, therefore its results are more photorealistic than DPST.}
\label{fig:results}
\end{figure}

\vspace*{-0.4cm}
\paragraph{Qualitative Assessment}
We start by providing a qualitative evaluation of our algorithm in Figure~\ref{fig:results}. Many more results are provided in the \href{http://cgm.technion.ac.il/Computer-Graphics-Multimedia/Software/photorealism/ID679-supp/index.html}{supplementary}, and we encourage the reader to view them at full size on a screen. 
%Two zoom examples are shown in Figure~\ref{fig:zoom}.
Several advantages of our method over DPST can be observed: 
(i) Fine details are better preserved. Our output images do not have smoothing-like effect, e.g. the rock crevices ($3^{rd}$ row).
%and the wood texture ($4^{th}$ row). 
(ii) Our method is better at preserving image boundaries, e.g., the billboards ($2^{nd}$ row) and the champagne bubbles ($4^{th}$ row).  
(iii) The identity of the content image is nicely preserved, as in the city scape and buildings ($1^{st}$ row). 
%These effects are also demonstrated in Figure~\ref{fig:zoom} which presents zoomed in examples.

\begin{figure}[t]
		\centering
        \setlength{\tabcolsep}{.15em}
        \scriptsize
        \begin{tabular}{c|ccc} 
        \includegraphics[width=.24\linewidth]{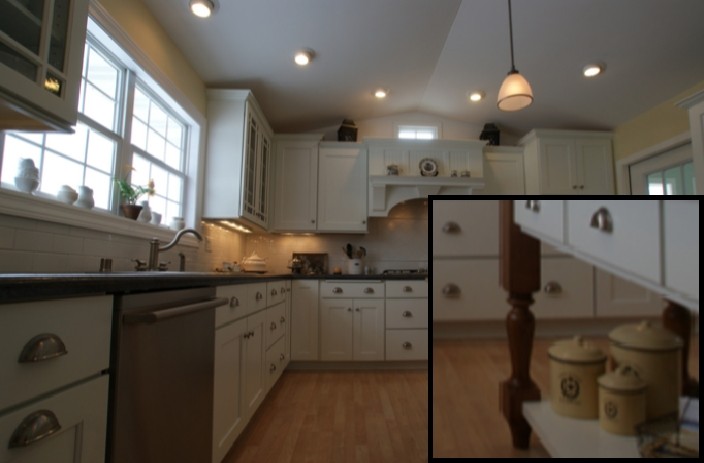}&
		\includegraphics[width=.24\linewidth]{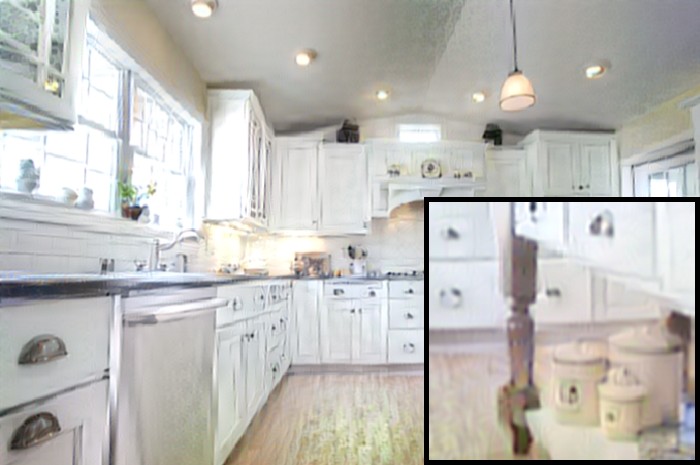}&
		\includegraphics[width=.24\linewidth]{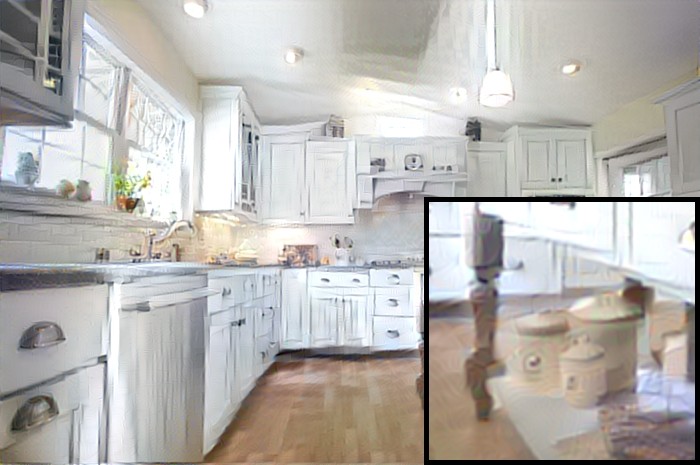}&
		\includegraphics[width=.24\linewidth]{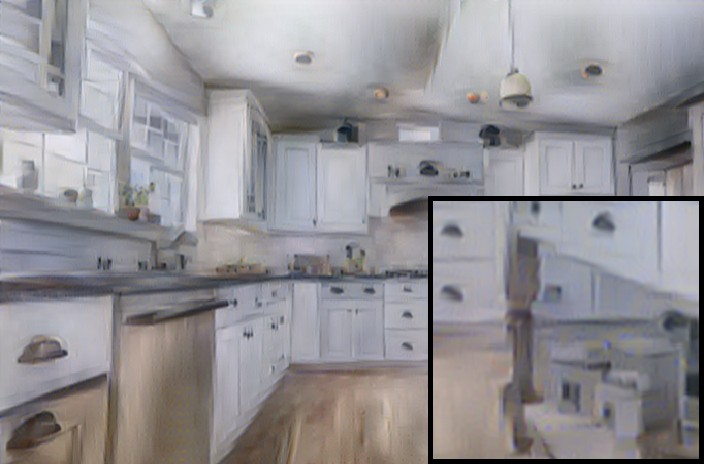}\\
        Input image& NS& NS+segment& CNNMRF \\
        \includegraphics[width=.24\linewidth]{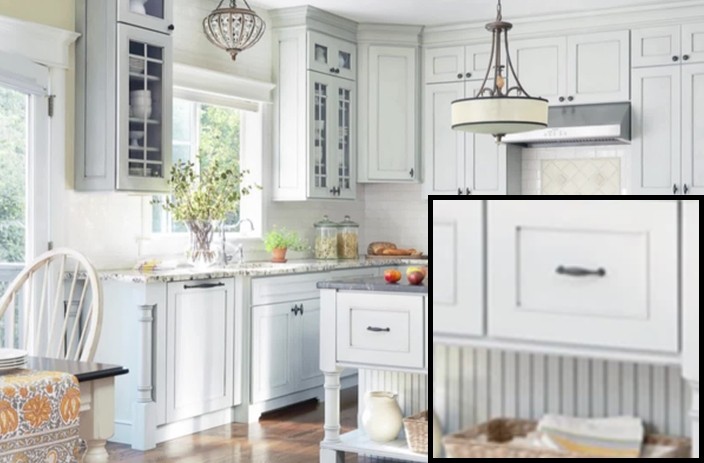}&
		\includegraphics[width=.24\linewidth]{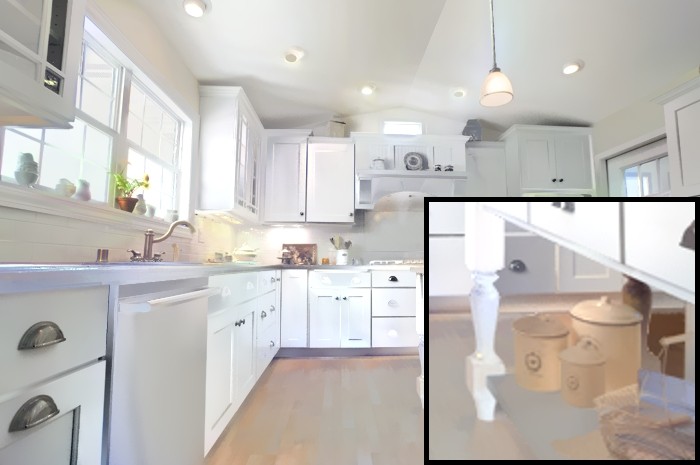}&
		\includegraphics[width=.24\linewidth]{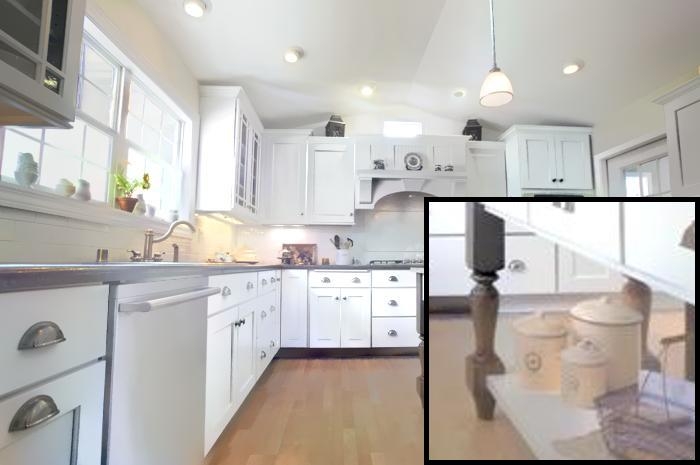}&
		\includegraphics[width=.24\linewidth]{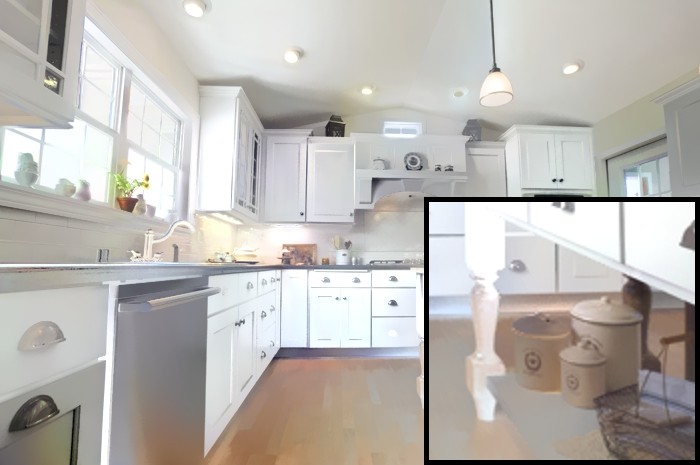}\\
        Style image& NS + \textbf{ML}& NS+segment + \textbf{ML} & CNNMRF + \textbf{ML} \\
        %& with ML& with ML& with ML\\
		%\includegraphics[width=.24\linewidth]{tar40.jpg}}
        &
		\includegraphics[width=.24\linewidth]{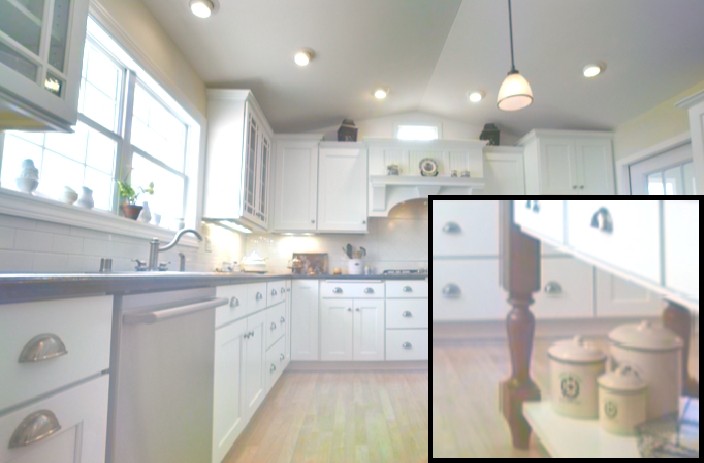}&
		\includegraphics[width=.24\linewidth]{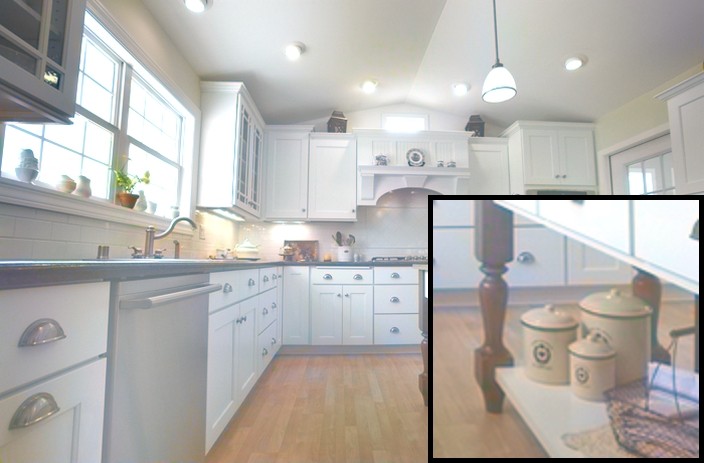}&
		\includegraphics[width=.24\linewidth]{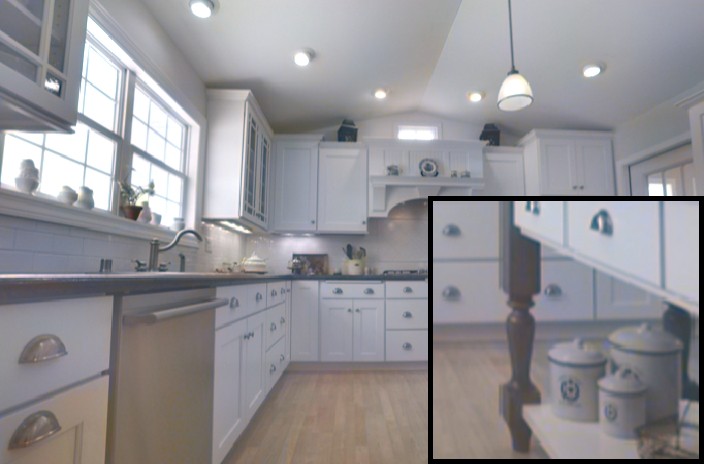}\\
        & NS + \textbf{SPE}& NS+segment + \textbf{SPE} & CNNMRF + \textbf{SPE} \\
        %& with SPE& with SPE& with SPE\\
        \end{tabular}
		\caption{\textbf{Generality to stylization method:} Our SPE can be combined with any style transfer method and will improve its photorealism. We compare SPE to the Matting Laplacian (ML) regularization of~\cite{luan2017deep}. It can be observed that SPE effectively restores the image gradients and yields photorealistic images while preserving the desired style. ML results often suffer from painting-like look.
        }
        %\vspace{-0.5cm}
		\label{fig:o}
\end{figure}

Figure~\ref{fig:o} shows the use of our SPE solver combined with other style transfer algorithms, this result makes our SPE solution quite \textit{general} for the task of enforcing gradient constraint on a stylized images. When comparing between ML and SPE post-processing, SPE preserves more fine details than ML using any of the three style transfer methods. 

Another comparison we provide is to methods for color transfer. 
Numerous such methods have been proposed. 
In Figure~\ref{fig:compare} we compare to three representatives.
It can be observed that our method is better in transferring the style (e.g., the lake and sky colors) while being less prone to artifacts due to lack of semantic information.

%\input{fig_compare_wu.tex}

%Figure~\ref{fig:plus} is an additional compare with the regulizer term based on the matting laplacian as suggested in \cite{luan2017deep}. In this example the matting laplacian is combined with the CNNMRF method \cite{li2016combining} and help in enforcing the realism on the image. Similar to the results of combining the matting laplacian with NS (i.e. DSPT) the results are tends to have highly homogeneous regions and lack of fine details -- this properties impair the photorealism of the outcome.  In contrast, our method transfer the original gradients including the fine details and textures (see the crops) and by that lead to a more photo-realistic outcome. 
%\input{fig_plus.tex}

%==========================================
\vspace*{-0.4cm}
\paragraph{Computation Time}
Our framework consists of two steps, the style transfer and the SPE post-processing. Our SPE takes 1.7sec for $640 \times 400$ images using MATLAB least square solver for each channel in parallel. A fast option for the stylization is StyleSwap with a pre-trained inversion net that takes 1.25sec. The overall time is thus less than 3sec. 
In comparison, Luan et al.~\cite{luan2017deep} report 3\textasciitilde5 minutes, (on the same GPU). To add to this, pre-processing of a matting matrix is needed as well as semantic segmentation computation of the input and style images. The ML post-processing alone takes 2\textasciitilde3 minutes.
This makes our approach much more attractive in terms of speed.

%==========================================
\vspace*{-0.4cm}
\paragraph{User Survey:}
%\textbf{Realism}
To assess our success in photorealistic style transfer we ran two user surveys. Both surveys were performed on a data-set of 40 images taken from~\cite{luan2017deep} (excluding the unrealistic input images). These surveys are similar to the ones suggested in~\cite{luan2017deep}, however, they used only 8 images.

The first survey assesses realism. Each image was presented to human participants who were asked a simple question: ``Does the image look realistic?''. The scores were given on a scale of [1-4], where 4 is 'definitely realistic' and 1 is 'definitely unrealistic'. We used Amazon Mechanical Turk (AMT) to collect 30 annotations per image, where each worker viewed only one version of each image out of three (DPST, our SPE and the original input image). 

To verify that participants were competent at this task, the average score they gave the original set of images was also recorded. We assume that the original images are 'definitely realistic', hence, workers that gave average score lower than 3 ('realistic') were excluded from the survey. For reference, the total average score, over all users, given to the original images is 3.51.

The second survey assesses style faithfulness.
Corresponding pairs of style and output images were presented to human participants who were asked a simple question: ``Do these two images have the same style?''. The scores were given on a scale of [1-4], where 4 is 'definitely same style' and 1 is 'definitely different style'. Again, we used AMT to collect 30 annotations per image, where each worker viewed either our result or that of DPST.  

The nature of this question is subjective and it is hard to validate the reliability of the survey by checking the participants competency, yet we argue that 30 annotations make the average results meaningful. In fact, we have noticed that the average scores become stable once we use 15 or more annotations.

Figure~\ref{fig:realism} plots the fraction of images with average score larger than a (i) photorealism and (ii) style-faithfulness score $\in [1,4]$. The overall Area-Under-Curve (AUC) values appear in the table.
Combining SPE with either StyleSwap or NeuralStyle leads to more photorealistic results than DPST, with an improvement of $>18\%$. 
In terms of style faithfulness, NeuralStyle+SPE results are similar to DPST, and both outperform StyleSwap+SPE. One should note, however, the significant runtime advantage of StyleSwap.

\definecolor{a}{rgb}{0.3176,0.5882,1.0000}
\definecolor{b}{rgb}{0.7294,0,0}
\definecolor{c}{rgb}{0.8392,0.8863,0.4784}
\definecolor{d}{rgb}{0.7,0.7,0.7}

\begin{figure}[t]
\centering
\tabcolsep=0.13cm
\small
\begin{tabular}{ccc}
\bmvaHangBox{{\includegraphics[width=.3\linewidth]{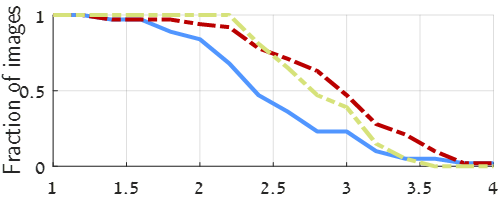}}}&
\bmvaHangBox{{\includegraphics[width=.3\linewidth]{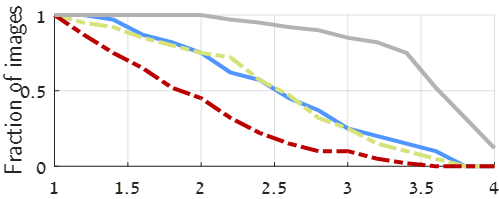}}}&
\bmvaHangBox{{
\scalebox{0.65}{
\tabcolsep=0.13cm
\begin{tabular}{ l | c | c }
 & Realism& Style \\
Method & score & score \\
\hline
\cellcolor{c!65}{NS+SPE}   		& \textbf{0.53} & \textbf{0.63} \\
\cellcolor{a!65}{Swap+SPE}   	& \textbf{0.54} & 0.52 \\
\cellcolor{b!65}{DPST}   		& 0.35 & \textbf{0.66} \\
\cellcolor{d!65}{original}    	& 0.87 & --- \\
\end{tabular}}
}}\\
(a) Style faithfulness & (b) Photorealism &(c) AUC values
\end{tabular}
\caption{\textbf{Quantitative evaluation.} Realism and style faithfulness scores obtained via a user survey (see text for details). The curves show the fraction of images with average score greater than \emph{score}. The Area-Under-Curve (AUC) values are presented in the table on the right.
StyleSwap+SPE and NeuralStyle+SPE provide significantly higher photorealism scores than DPST.
NeuralStyle+SPE and DPST outperform StyleSwap in terms of style faithfulness.}
  \label{fig:realism}
\end{figure}

\begin{figure}
		\setlength{\tabcolsep}{.19em}
        \small
        \begin{tabular}{cc|ccc} 
		\includegraphics[width=.19\linewidth]{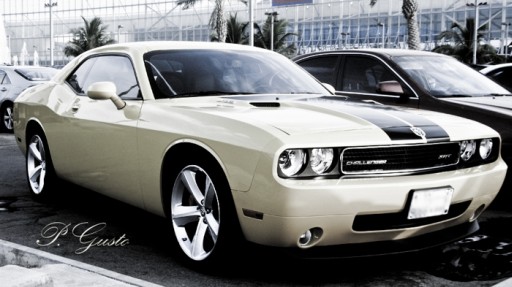}&
		\includegraphics[width=.19\linewidth]{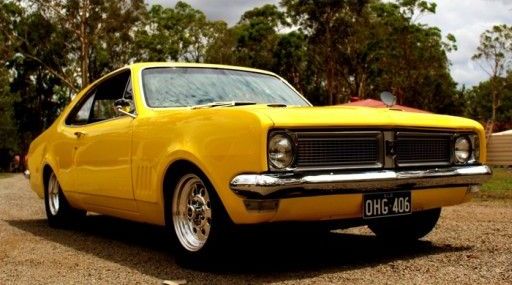}&
		\includegraphics[width=.19\linewidth]{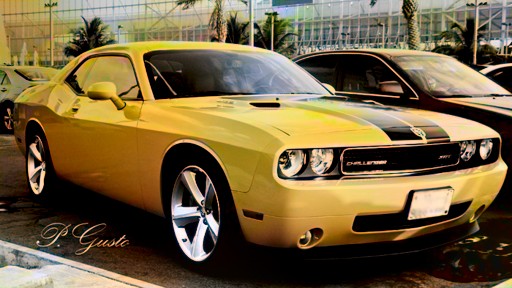}&
		\includegraphics[width=.19\linewidth]{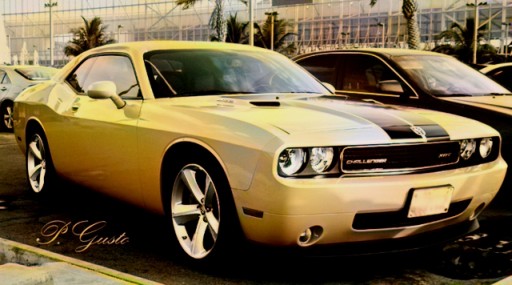}&
		\includegraphics[width=.19\linewidth]{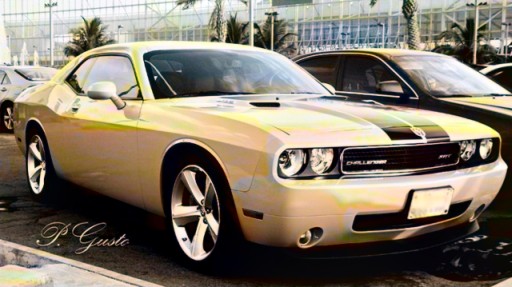}\\
        \includegraphics[width=.19\linewidth]{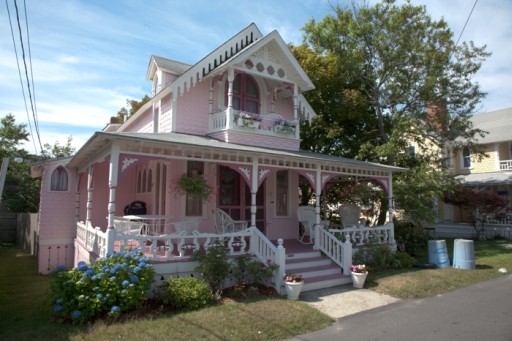}&
		\includegraphics[width=.19\linewidth]{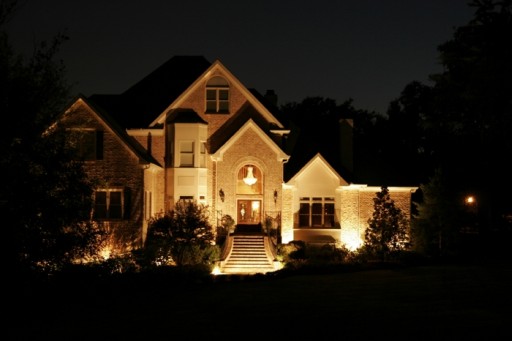}&
		\includegraphics[width=.19\linewidth]{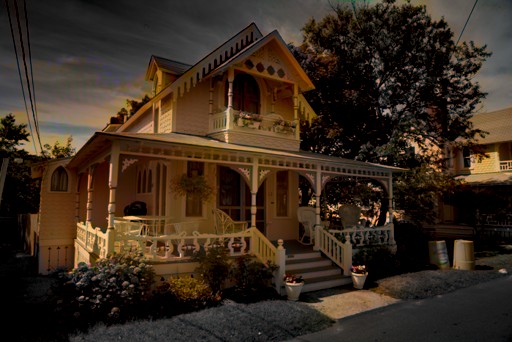}&
		\includegraphics[width=.19\linewidth]{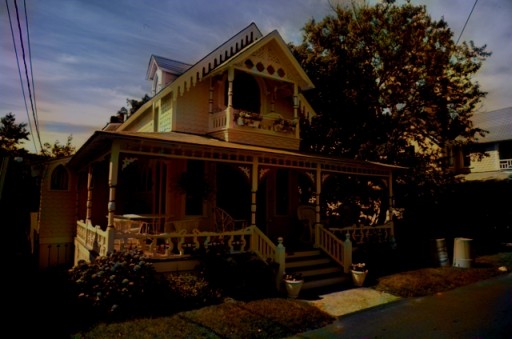}&
		\includegraphics[width=.19\linewidth]{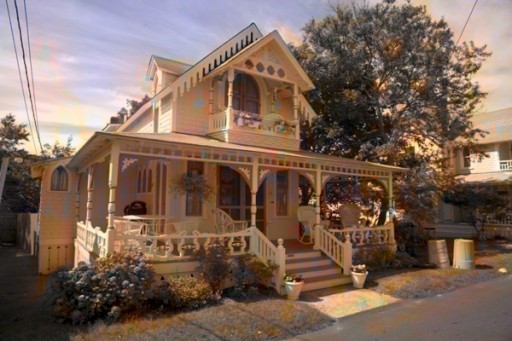}\\
        \includegraphics[width=.19\linewidth]{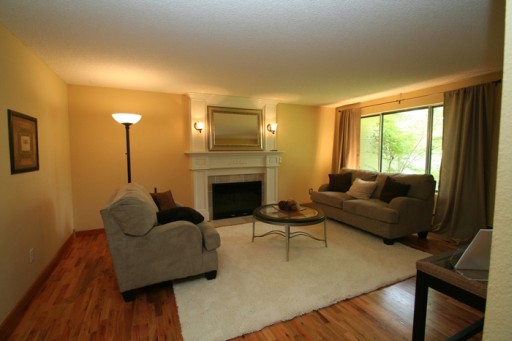}&
		\includegraphics[width=.19\linewidth]{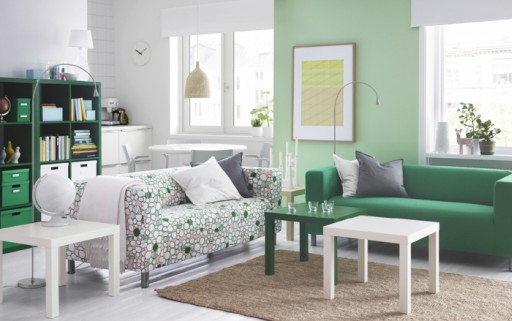}&
		\includegraphics[width=.19\linewidth]{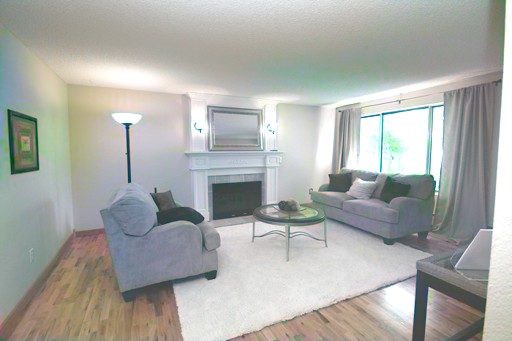}&
		\includegraphics[width=.19\linewidth]{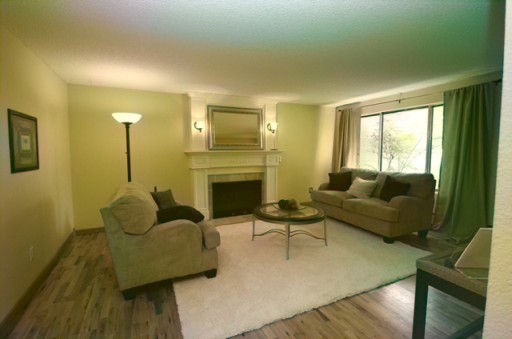}&
		\includegraphics[width=.19\linewidth]{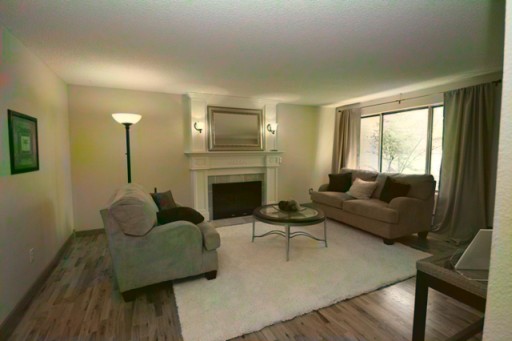}\\
        Input image & Style image & NS+SPE & StyleSwap+SPE & CNNMRF+SPE
        \end{tabular}
		\caption{Balance between speed and style faithfulness. SPE can be combined with any style transfer method. Combining SPE with NeuralStyle or CNNMRF yields impressive realistic stylization results, yet these methods' computation time is long. In contrast combining SPE with StyleSwap results in a very fast overall method, with weaker style transfer capabilities. }
		\label{fig:general}
\end{figure}

Our empirical evaluation and the user survey suggest that StyleSwap transfers the style significantly less successfully than NS (or DPST). On the other hand, while DPST achieved impressive realistic stylization results it is prone to artifacts and is very slow. We suggest a simple alternative that has much less artifacts, better maintains high frequencies and transfers style in a similar way. Our method can be combined with fast stylization methods like StyleSwap, that does not require segmentation, providing an overall fast solution, but weaker in terms of transferring style. 
This balance between speed and the ability to transfer the style is demonstrated in Figure~\ref{fig:general}.
As fast stylization methods get better our method will become better too.

\section{Conclusions and Limitations}
\vspace*{-0.2cm}
\label{sec:conc}
We propose a two-stage framework for photorealistic style transfer. Our main contribution is the observation that a simple post-processing, based on the Screened Poisson Equation, can significantly improve the photorealism of a stylized image. One limitation of SPE with respect to DPST is dealing with significant contrast inversion (e.g., second and sixth rows in Fig. \ref{fig:results}), SPE tends to create 'halo' artifacts around strong edges. 
As it appears, end-to-end learning of photorealism is still an open question, hence, post-processing methods like the one we suggest, are currently the most viable option for deep style transfer.
\\
%Also, like in \cite{luan2017deep}, semantic segmentation is infrequently needed to avoid content bleeding artifacts, however methods for semantic segmentation only work well for common labels (e.g., sky, grass, buildings). 

\noindent\textbf{Acknowledgements}
This research was supported by the Israel Science Foundation under Grant 1089/16, by the Ollendorf foundation and by Adobe. We thank Itamar Talmi for his useful comments and helpful discussions.

\clearpage

\bibliography{egbib}
\end{document}